%% file: berndt2020tro.tex
\documentclass[journal]{IEEEtran}

\usepackage{cite}
\usepackage{graphics}    
\usepackage{times}       
\usepackage{mathrsfs}
\usepackage{url}

\usepackage{amsmath}     
\usepackage{amssymb}     
\usepackage{graphicx}
\usepackage{color}
\usepackage[noend]{algorithmic}
\usepackage{algorithm}
\usepackage{xspace} 

\input{utils/packages.tex}

\input{utils/glossary.tex}

\usepackage[font={small}]{caption}

\input{utils/theorems.tex}
\input{utils/refs.tex}

\input{utils/symbols.tex}
\input{utils/devtools.tex}

\title{\LARGE \bf Receding Horizon Re-ordering of Multi-Agent 
  Execution Schedules} 

\input{utils/authors.tex}
\begin{document}

\maketitle

%
\input{sec/0_abstract.tex}
\IEEEpeerreviewmaketitle
\input{sec/1_intro.tex}

\input{sec/2_related.tex}
\input{sec/3_adg.tex}
\input{sec/4_reordering.tex}

\input{sec/5_ocp_feedback.tex}

\input{sec/6_rhc.tex}

\input{sec/7_alternative_solvers.tex}
\input{sec/8_eval.tex}
\input{sec/9_conclusion.tex}

\section*{Acknowledgments}

The authors would like to thank Musa~Morena~Marcusso~Manh{\~a}es
  for her help with setting up the \acs{ROS} and Gazebo frameworks
  as well as Zhe Chen for the help setting up the code to compare 
  our approach with K-CHSH-RM solver.
  
\input{sec/a_appendix.tex}

\ifCLASSOPTIONcaptionsoff
  \newpage
\fi

\bibliographystyle{ieeetr}
\bibliography{bib/references}

\input{utils/biographies.tex}

\end{document}

%% file: utils/packages.tex
\usepackage[shortcuts,acronym]{glossaries}
 
\usepackage{amsthm}

\usepackage[T1]{fontenc}

\usepackage{multirow}
\usepackage{graphicx}
\usepackage{subcaption}
\usepackage{lipsum} 

\usepackage{tikz-cd}
\usetikzlibrary{shapes}
\usetikzlibrary{arrows.meta}
\usetikzlibrary{arrows}
\usetikzlibrary{positioning}
\usetikzlibrary{fadings}
\tikzset{
  shift left/.style ={commutative diagrams/shift left={#1}},
  shift right/.style={commutative diagrams/shift right={#1}}
}

\definecolor{colorborder}{RGB}{20,20,20} 
\definecolor{colorinside}{RGB}{240,240,240} 

%% file: utils/glossary.tex
\newacronym{ADG}{ADG}{Action Dependency Graph}
\newacronym{ADGs}{ADGs}{Action Dependency Graphs}
\newacronym{SE-ADG}{SE-ADG}{Spatially Exclusive Action Dependency Graph}
\newacronym{SE-ADGs}{SE-ADGs}{Spatially Exclusive Action Dependency Graphs}
\newacronym{SADG}{SADG}{Switchable Action Dependency Graph}
\newacronym{SADGs}{SADGs}{Switchable Action Dependency Graphs}
\newacronym{AGV}{AGV}{Automated Guided Vehicle}
\newacronym{AGVs}{AGVs}{Automated Guided Vehicles}
\newacronym{AMR}{AMR}{Autonomous Mobile Robot}
\newacronym{AMRs}{AMRs}{Autonomous Mobile Robots}
\newacronym{CBC}{CBC}{Coin-Or Branch-and-Cut}
\newacronym{CBS}{CBS}{Conflict-Based Search}
\newacronym{CCBS}{CCBS}{Continuous Conflict-Based Search}
\newacronym{DMAPF}{D-MAPF}{Dynamic Multi-Agent Path Finding}
\newacronym{ECBS}{ECBS}{Enhanced Conflict-Based Search}
\newacronym{MAPF}{MAPF}{Multi-Agent Path Finding}
\newacronym{MAPD}{MAPD}{Multi-Agent Pickup and Delivery}
\newacronym{MILP}{MILP}{Mixed-Integer Linear Program}
\newacronym{MILPs}{MILPs}{Mixed-Integer Linear Programs}
\newacronym{OCP}{OCP}{Optimal Control Problem}
\newacronym{UAV}{UAV}{Unmanned Aerial Vehicle}
\newacronym{RHC}{RHC}{Receding Horizon Control}
\newacronym{SHC}{SHC}{Shrinking Horizon Control}
\newacronym{ROS}{ROS}{Robot Operation System}
\newacronym{SIPP}{SIPP}{Safe Interval Path Planning}

%% file: utils/theorems.tex
\newtheoremstyle{mainstyle}
  {\topsep} 
  {\topsep} 
  {\itshape} 
  {} 
  {\bfseries \itshape} 
  {.} 
  {.5em} 
  {} 
\newtheoremstyle{proofstyle}
  {\topsep} 
  {\topsep} 
  {\itshape} 
  {} 
  {\itshape} 
  {:} 
  {.5em} 
  {} 

\theoremstyle{mainstyle}
\newtheorem{myassumption}{Assumption} \def\assref#1{Assumption~\ref{#1}}
 
\newtheorem{mydefinition}{Definition} \def\defref#1{Definition~\ref{#1}}
\newtheorem{mylemma}{Lemma} \def\lemref#1{Lemma~\ref{#1}}
\newtheorem{myproposition}{Proposition} 
\newtheorem{mycorollary}{Corollary} 
\newtheorem{myremark}{Remark} 

\theoremstyle{proofstyle}
\newtheorem{myproof}{Proof}  

%% file: utils/refs.tex
\def\algref#1{Algorithm~\ref{#1}}
\def\secref#1{Section~\ref{#1}}
\def\figref#1{Fig.~\ref{#1}}
\def\tabref#1{Table~\ref{#1}}
\def\corref#1{Corollary~\ref{#1}}
\def\eqref#1{(\ref{#1})}
\def\eqnref#1{(\ref{#1})}
\def\remref#1{Remark~\ref{#1}}

%% file: utils/symbols.tex
\newcommand{\status}[1]{\textit{status}(#1)}

\newcommand{\staged}{\emph{staged}}
\newcommand{\inprogress}{\emph{in-progress}}
\newcommand{\completed}{\emph{completed}}

\newcommand\SEADGgraph{\mathcal{G}_\text{\acs{SE-ADG}}}
\newcommand\SEADGvertices{\mathcal{V}_\text{\acs{SE-ADG}}}
\newcommand\SEADGedges{\mathcal{E}_\text{\acs{SE-ADG}}}
\newcommand\SEADGgraphset{\mathbb{G}_\text{\acs{SE-ADG}}}

\newcommand\bvec{\textbf{\textit{b}}}
\newcommand\zerovec{\textbf{\textit{0}}}
\newcommand{\SADGgraph}[1]{\mathcal{G}_\text{\acs{SADG}}(#1)}        
\newcommand{\SADGvertices}[1]{\mathcal{V}_\text{\acs{SADG}}(#1)}     
\newcommand{\SADGedges}[1]{\mathcal{E}_\text{\acs{SADG}}(#1)}        

\newcommand\edgeFWD{e_\text{fwd}}
\newcommand\edgeREV{e_\text{rev}}

\newcommand\binaryMapFunc{\textit{depSwitch}}
\newcommand{\binaryMap}[3]{\binaryMapFunc\big(#1:\{ #2, #3 \} \big)} 

\newcommand{\optOCP}[1]{f_\text{OCP}(#1)}
\newcommand\bvecstart{\bvec_0}

\newcommand\tsvec{\textbf{\textit{t}}_s}
\newcommand\tgvec{\textbf{\textit{t}}_g}

\newcommand\bvecFH{\bvec_\text{rhc}}
\newcommand{\SADGgraphFH}[1]{\mathcal{G}^{rhc}_\text{\acs{SADG}}(#1)}  
\newcommand\horizon{H}
\newcommand\SADGverticesFH{\mathcal{V}_\text{FH}}
\newcommand\SADGedgesFH{\mathcal{E}_\text{FH}}

\newcommand{\SEADGgraphSubset}{\mathcal{G}^{rhc}_\text{\acs{SE-ADG}}}

\newcommand\ms{ \textit{ m}\cdot \textit{s}^{\scriptscriptstyle{-1}}} 
\newcommand\rads{ \textit{ rad}\cdot \textit{s}^{\scriptscriptstyle{-1}}} 
\newcommand\delaytime{\Delta t_\text{delay}}

\newcommand\Kalg{K-CBSH-RM\xspace}

%% file: utils/devtools.tex
\newcommand{\mychange}[1]{{#1}}

%% file: utils/authors.tex
\author{Alexander~Berndt$^{1}$, Niels~van~Duijkeren$^{2}$, Luigi~Palmieri$^{2}$, 
        Alexander~Kleiner$^{2}$, and Tam\'as~Keviczky$^{3}$ 
	\thanks{The research that lead to this paper was funded by Robert Bosch GmbH. 
			This work was supported by the European Union's Horizon 2020 
			research and innovation program 
			under grant agreement No. 101017274 (DARKO).}
	\thanks{$^{1}$Alexander~Berndt is with 
	  Overstory B.V., 1018 VN Amsterdam, The Netherlands
	  		{\tt\small berndtae@gmail.com }}%
	\thanks{$^{2}$Niels~van~Duijkeren, Luigi~Palmieri and Alexander~Kleiner are 
		with Robert Bosch GmbH, Corporate Research,
		Renningen, 71272, Germany
		{\tt\small \{ Niels.vanDuijkeren, Luigi.Palmieri, Alexander.Kleiner \} 
		@de.bosch.com}}%
	\thanks{$^{3}$Tam\'as~Keviczky is with 
      the Delft Center for Systems and Control (DCSC),
      TU Delft, 2628 CN Delft, The Netherlands
            {\tt\small T.Keviczky@tudelft.nl }}%
	
}

%% file: sec/0_abstract.tex
\begin{abstract}
    The trajectory planning for a fleet of \acf{AGVs} on a roadmap is  
      commonly referred to as the \ac{MAPF} problem,
      the solution to which dictates each \acs{AGV}'s spatial and temporal 
      location until it reaches its goal without collision.
    When executing \acs{MAPF} plans in dynamic workspaces,
      \acs{AGVs} can be frequently delayed, e.g.,
      due to encounters with humans or third-party vehicles.
    If the remainder of the \acs{AGVs} keeps following their individual plans,
      synchrony of the fleet is lost and some \acs{AGVs} may pass 
      through roadmap intersections
      in a different order than originally planned.
    Although this could reduce the cumulative route completion 
      time of the \acs{AGVs}, generally,
      a change in the original ordering can 
      cause conflicts such as deadlocks.
    In practice,
      synchrony is therefore often enforced by 
      using a \acs{MAPF} execution policy employing, e.g., 
      an \ac{ADG} to maintain ordering. 
    To safely re-order without introducing deadlocks,
      we present the concept of the \acf{SADG}.
    Using the \acs{SADG}, we formulate a comparatively low-dimensional \acf{MILP} that 
      repeatedly re-orders \acs{AGVs} in a recursively feasible manner,
      thus maintaining deadlock-free guarantees,
      while dynamically minimizing the cumulative route completion time of all \acs{AGVs}.
    Various simulations validate the efficiency of our approach
      when compared
      to the original \acs{ADG} method as well as robust \acs{MAPF} solution approaches.
  \end{abstract}
  
  \begin{IEEEkeywords}
    Robust Plan Execution, 
    Scheduling and Coordination, 
    Mixed Integer Programming, 
    Multi-Agent Path Finding.
  \end{IEEEkeywords}

%% file: sec/1_intro.tex

\vspace{-3mm}
\section{Introduction}
\label{sec:introduction}


\IEEEPARstart{M}{ultiple} \acf{AMRs} have been shown to significantly 
  increase the efficiency of performing intralogistics tasks such as moving 
  inventory in distribution centers 
  \cite{wurmanCoordinatingHundredsCooperative2008}.
Coordinating \acs{AMRs} navigating a shared environment 
  can be formulated as the \acf{MAPF} 
  problem \cite{sternMultiAgentPathfindingDefinitions2019}.
The \acs{MAPF} problem is to find trajectories for each \acs{AMR}
  along a roadmap such that 
  each \acs{AMR} reaches its goal without colliding with the others, 
  while minimizing a cost metric such as 
  the makespan or cumulative route completion time (also referred to as sum-of-costs).
Throughout this manuscript, 
  we refer to \acs{AMRs} as \acf{AGVs} to be consistent 
  with the \acs{MAPF} literature.  

\begin{figure}[t!]
	\centering
	\begin{subfigure}{0.48\textwidth}
		\centering
		\includegraphics[width=1.0\linewidth]{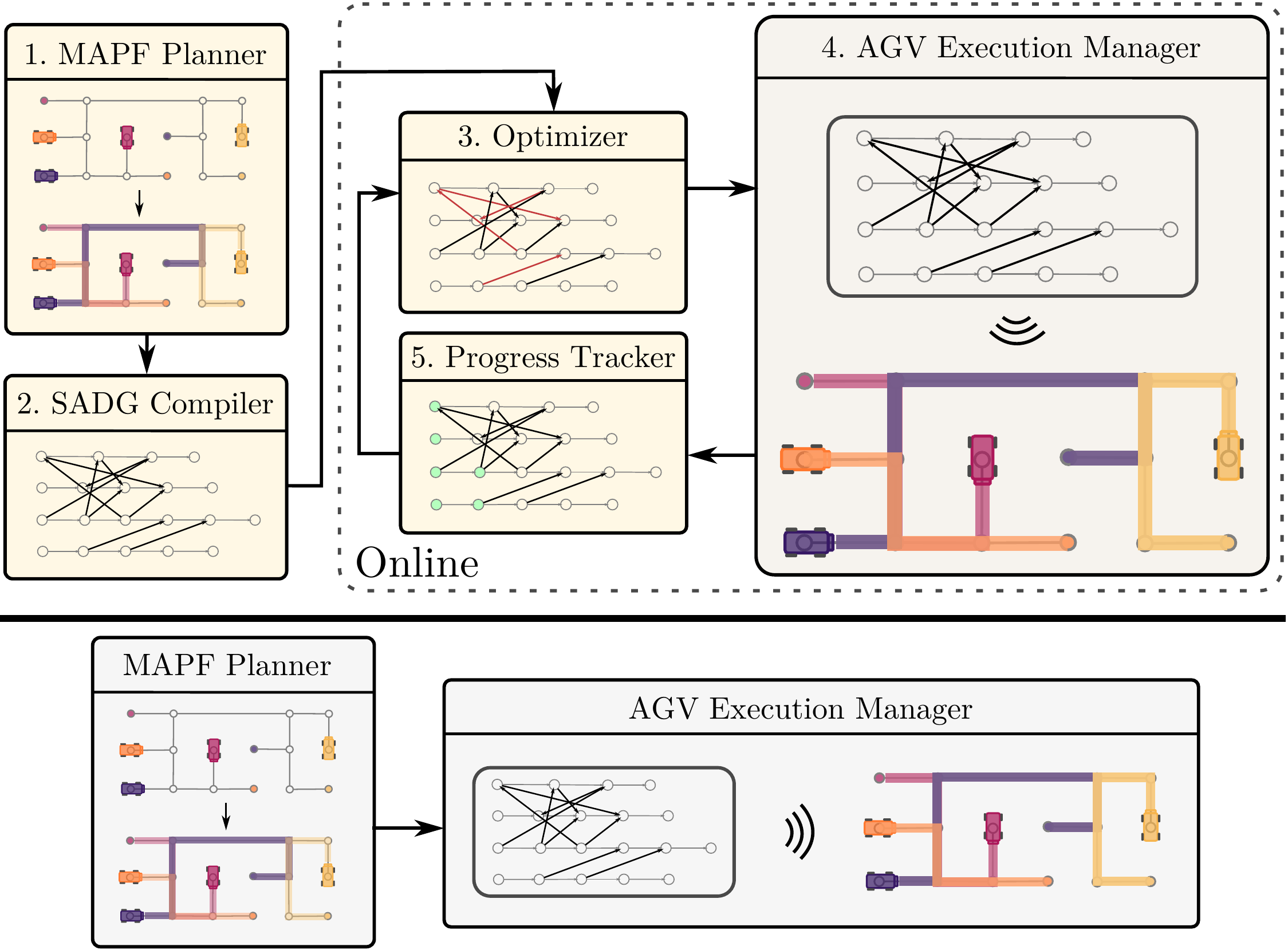}
	\end{subfigure}
  \vspace{3mm}
	\caption{
    \textbf{Top:} Our proposed optimization-based feedback control scheme. 
    \textbf{Bottom:} Typical \acs{MAPF} plan execution schemes.
    Our approach significantly reduces the cumulative route completion 
      of \acs{AGVs} subjected to large delays by optimizing the 
      ordering of \acs{AGVs} based on their progress in a receding 
    horizon fashion,
    while maintaining collision- and deadlock-free plan execution guarantees.
  }
	\vspace{-3mm}
  \label{fig:covergirl}
\end{figure}  

\mychange{
Minimizing temporal cost metrics when solving the \acs{MAPF} problem 
  has received a lot of attention 
  in the literature \cite{yuMultiagentPathPlanning2013}.
}
However, even if an optimal \acs{MAPF} solution is found, 
  blindly executing the plans can still result in deadlocks when the \acs{AGVs}
  experience delays.
This introduces the need for plan execution policies,
  used to maintain the ordering between \acs{AGVs}
  and thus avoiding deadlocks.
The authors in \cite{hoenigPersistentRobustExecution2018} 
  propose compiling an \acf{ADG} from a \acs{MAPF} solution
  to enforce the ordering during plan execution. 
However, this work, and most other works 
  such as \cite{atzmonRobustMultiAgentPath2020,
  hoenigMultiAgentPathFinding2016,
  maMultiAgentPathFinding2017} 
  consider \acs{AGVs} which are only marginally delayed.
This means that delays are seen as a lack of synchronization
  between \acs{AGVs},
  rather than significantly affecting the overall route completion times.
With the advent of Industry 4.0 (such as the VDA5050 
  protocol \cite{vda5050} and the Robot Middleware Framework \cite{rmf_framework}),
  we turn our attention to \acs{AGV} fleets navigating 
  dynamic and complex environments occupied by humans
  and third-party vehicles.
These dynamic environments are far less predictable than
  those typically considered in the \acs{MAPF} literature,
  implying that \acs{AGVs} can experience large delays when waiting for, e.g., 
  a human to move out of its path.

These large, unpredictable delays can result in 
  inefficient plan execution because the implicit ordering of the original 
  \acs{MAPF} solution requires \acs{AGVs} to wait for largely delayed \acs{AGVs}.
Not adhering to this implicit ordering, however, can result in deadlocks.
\mychange{Approaches like \cite{Cap2016, Coskun2021} propose re-ordering schemes
  which maintain the deadlock-freeness properties of the original plan.
The decision to switch the order between two robots or not, however, 
  is based only on performance measures of the two involved \acs{AGVs}.
This means that although these switches are performed throughout the fleet,
  they do not necessarily lead to an overall performance increase in terms of 
  a sum-of-costs or a makespan metric.
}

\paragraph*{Contributions}
\mychange{
To address the shortcomings of robust \acs{MAPF} approaches 
  and local path repair methods, we present:
1.) The \acf{SADG}, a novel data structure which  
    formalizes the definition of switchable 
    dependencies between \acs{AGVs} in multi-agent plans;
2.) An online optimization-based 
    \acf{SHC} scheme which re-orders \acs{AGVs}
    based on the \acs{AGV}s' current progress 
    along their paths 
    while maintaining collision- and deadlock-freeness guarantees;
3.) Extension of the \acs{SHC} to a \acf{RHC} scheme,
    which significantly reduces computation times and thereby 
    enables real-time applications for
    all of the presented maps and team 
    sizes without sacrificing collision- and deadlock-freeness guarantees.
We compare our approach to the baseline \acs{ADG} method presented in 
  \cite{hoenigPersistentRobustExecution2018} 
  as well as the state-of-the-art robust \acs{MAPF} solver \Kalg{} 
  \cite{atzmonRobustMultiAgentPath2020}, 
  yielding up to a $25\%$ overall decrease 
  in average route completion times as robots are confronted with large delays.
Our method is available online as an open-source software 
  package called \textit{sadg-controller}\footnote{\url{https://github.com/alexberndt/sadg-controller}}. 
}
This manuscript extends ideas and concepts detailed in a preliminary 
  workshop paper presented at the 30th International 
  Conference on Automated Planning and Scheduling (ICAPS), 
  Nancy, France, October 2020 \cite{berndtFeedbackSchemeSADGICAPS2020}.

\paragraph*{Outline}
\secref{sec:related_work} 
  presents existing solutions and their 
  capabilities and shortcomings in the
  context of our proposed solution.
Preliminaries regarding the routing of multiple \acs{AGVs},
  as well as the problem formulation,
  is presented in \secref{sec:adg}.
We present the concept of the \acl{SADG} in \secref{sec:sadg}
  and formulate the mixed-integer \acs{OCP} in \secref{sec:shc}.
The method is extended to a receding horizon feedback control scheme 
  in \secref{sec:rhc}.
We evaluate our approach in \secref{sec:eval} and conclude the paper 
  in \secref{sec:conclusion}.



%% file: sec/2_related.tex

\vspace{-3mm}
\section{Related Work}
\label{sec:related_work}

Recently, solving the \acs{MAPF} problem has garnered 
  wide-spread attention 
  \cite{sternMultiAgentPathfindingDefinitions2019,felnerSearchBasedOptimalSolvers2017}.
This is mostly due to the abundance of application domains, 
  such as intralogistics, 
  airport taxi scheduling \cite{morris2016planning} 
  and computer games \cite{mOntanon2013survey}. 
Solutions to the \acs{MAPF} problem include 
  \ac{CBS} \cite{sharonConflictbasedSearchOptimal2015}, 
  Prioritized Planning \cite{kleiner2019prioritzedplanning},  
  declarative optimization approaches using answer 
  set programming \cite{bogatarkanDeclarativeMethodDynamic2019}, 
  heuristic-guided coordination \cite{pecora2018loosely} 
  and graph-flow optimization approaches \cite{yuPlanningoptimalpaths2013}.

Algorithms such as \acs{CBS} have been improved 
  by exploiting properties such as 
  geometric symmetry \cite{liSymmetryBreakingConstraintsGridBased2019}, or
  using purpose-built heuristics \cite{felnerAddingHeuristicsConflictBased2018}.
In \cite{lamBranchandCutandPriceMultiAgentPathfinding2019}, the authors
  reformulate the \ac{MAPF} as a \ac{MILP} and solve it 
  using a branch-cut-and-price approach.
\acs{MILP} formulations have also been used in numerous 
  binary-decision based receding horizon control problems,
  referred to as hybrid control systems \cite{bemporadMPChybrid2002}.  
Practical applications using these 
  formulations include 
  coordinating agents in urban road networks \cite{linRoadNetworkMILP2011},
  coordinating autonomous cars at intersections \cite{Hult2015,Ravikumar2021},
  \acs{UAV} trajectory planning \cite{richardsMILPtrajectoryOpt2002},
  multi-agent persistent coverage \cite{maria2022milppersistentcoverage}
  and train scheduling \cite{trainSchedulingMILP}.
Similarly, the development of 
  bounded sub-optimal solvers such as 
  \ac{ECBS} \cite{barerSuboptimalVariantsConflictBased2014} 
  have further improved planning performance for higher dimensional state spaces. 
In turn, 
  \ac{CCBS} extends \acs{CBS} by enabling planning on  
  roadmap graphs with weighted edges and considering continuous 
  time intervals to describe collision avoidance constraints, 
  albeit with increased solution times \cite{andreychukMultiAgentPathfindingContinuous2019}.

The abstraction of the \acs{MAPF} to a graph search problem requires simplifying
  assumptions to manage complexity.
These assumptions include the use of very crude vehicle motion models and
  neglecting most of the effects of unpredictable delays in stochastic and dynamic environments.
In order to maintain validity of the \acs{MAPF} plan during execution,
  it is required to synchronize the progress of all \acs{AGVs} by closely monitoring the fleet.
This synchronization can be achieved using a so-called 
  execution policy to manage the \acs{AGVs} according to their individual plans.

An \acf{ADG} encodes the ordering between \acs{AGVs} 
  as well as their kinematic constraints in a 
  post-processing step after solving the \acs{MAPF} 
  \cite{hoenigMultiAgentPathFinding2016}.
Combined with a plan execution policy, 
  this allows \acs{AGVs} to execute \acs{MAPF} plans successfully despite 
  kinematic constraints and unforeseen delays.
\mychange{
Closely related to the \acs{ADG}-based execution policy to account for disturbances
  is RMTRACK \cite{Cap2016}.
For every pair of robots,
  RMTRACK identifies collision regions (i.e., relative delays that lead to collisions) in their coordination space.
It is imposed that the trajectory in the coordination space remains homotopic to the undisturbed trajectory.
This leads to an equivalent coordination approach as \cite{hoenigMultiAgentPathFinding2016} and guarantees deadlock-freeness.
Follow-up work \cite{Coskun2019} proposes to relax the homotopy equivalence condition and allow flipping the order in which robots
  pass through a certain region considering two different optimization strategies.
The latter approach was later extended to guarantee deadlock-free plan execution \cite{Coskun2021}
  by asserting that the so-called ``segment graph'' that results from flipping the order has no cycles.
Note that the ``segment graph'' is closely related to the \acs{ADG} \cite{hoenigMultiAgentPathFinding2016}.
Moreover,
  the concept that flipping the order of two robots results in a different ``segment graph''  in turn relates to the
  \acs{SADG} presented in \cite{berndtFeedbackSchemeSADGICAPS2020} and this paper.
Contrary to these works,
  the authors of \cite{Coskun2021} do not consider changing the order of robots at every possible conflict simultaneously.
They thereby avoid the need to solve a costly \acs{MILP},
  but sacrifice optimality of the overall plan execution.
}
Since the \acs{MAPF} considers a fleet of \acs{AGVs} each with 
  a unique start and goal position,
  an additional framework is required to 
  allow for the persistent planning of \acs{AGVs}.
Such a framework is proposed in \cite{hoenigPersistentRobustExecution2018},
  where the aforementioned \acs{ADG} can be used to 
  anticipate where \acs{AGVs} will be in a future time-step (called a \textit{commit}), 
  allowing the \acs{MAPF} to be solved from there, 
  while the \acs{AGVs} execute the plans up until this \textit{commit}.


Several \acs{MAPF} methods have been introduced to particularly handle delays.
$k$R-\acs{MAPF} solvers such as \Kalg address this by permitting delays 
  up to a duration of $k$ time-steps 
  \cite{atzmonRobustMultiAgentPath2020, ChenAAAI21b}.
Stochastic \acs{AGV} delay distributions are considered in 
  \cite{maMultiAgentPathFinding2017}, 
  where the \acs{MAPF} is solved by minimizing the expected overall delay.
These robust \acs{MAPF} formulations and solutions 
  inevitably result in more conservative plans compared to their nominal counterparts.
A robust approach to 
  handle communication delays and packet losses
  for \acs{AGVs} with second-order dynamics is 
  considered in \cite{mannucci2019multirobotcomms}.
However, all these solutions   
  do not specifically address the effects of significantly large delays. 
These approaches typically view delays 
  as a bounded lack of synchronization between \acs{AGVs}, 
  rather than as significantly impacting the route completion time.

The contribution of this paper is a method that extends the concept of an \acs{ADG}
  by modeling the allowed re-orderings of \acs{AGVs} at intersections,
  obtaining an \acf{SADG}.
The routes for each \acs{AGV} are considered given and remain unaltered.
Typically they would be computed using an existing \acs{MAPF} solver.
The result is a comparatively low-dimensional decision-making problem,
  to continuously and reactively modify the \acs{MAPF} plan online 
  to improve the cumulative route completion time.
We formulate the problem as a \acs{MILP} that can be solved using 
  off-the-shelf---commercial as well as open-source---solvers.
By our re-ordering approach we allow \acs{AGVs} to continue with their tasks without 
  needing to unnecessarily wait for delayed \acs{AGVs}, 
  while guaranteeing deadlock- and collision-free execution.

%% file: sec/3_adg.tex
\begin{figure*}[h!]
	\centering
	\begin{subfigure}{0.48\textwidth}
		\centering
		\includegraphics[width=0.95\linewidth]{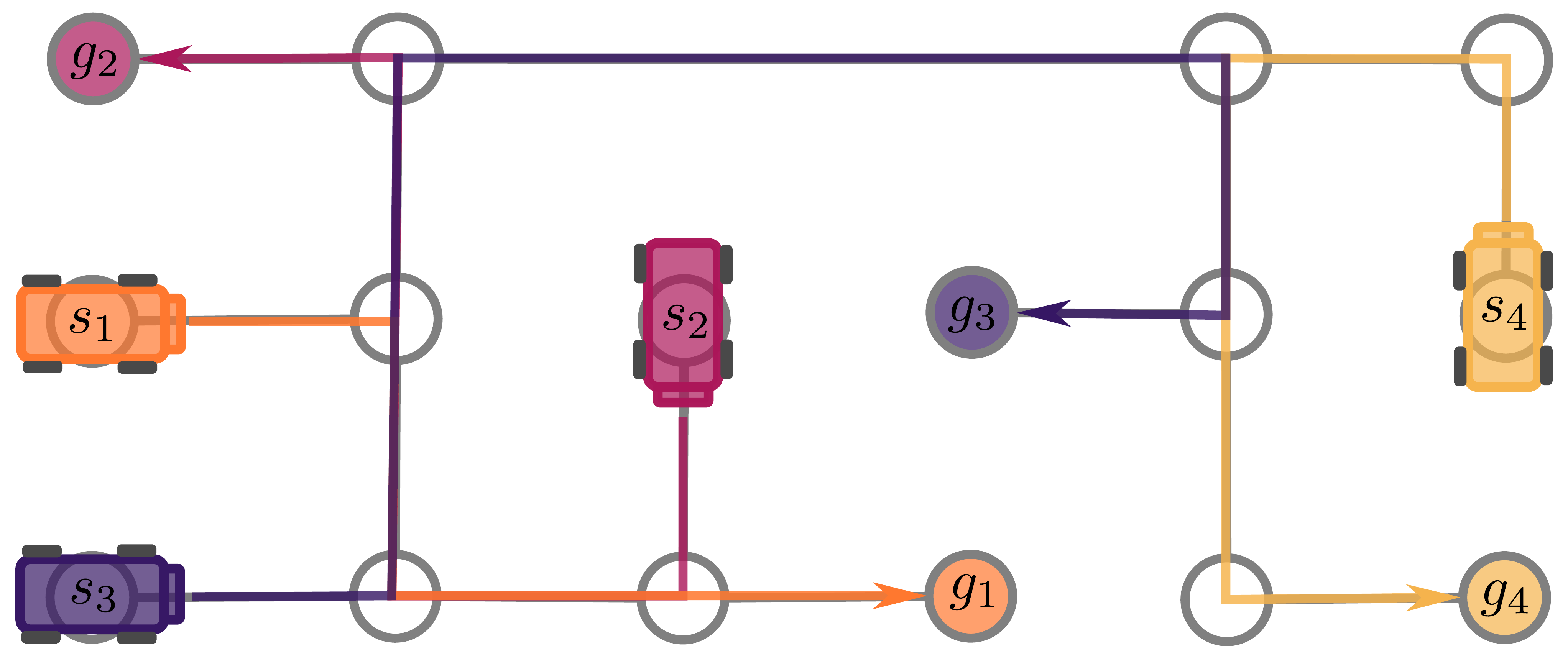}
		\caption{Roadmap $\mathcal{G} = (\mathcal{V},\mathcal{E})$ with $\mathcal{P}$}
		\label{fig:roadmap}
	\end{subfigure}
	\hspace{4mm}
	\begin{subfigure}{0.48\textwidth}
		\centering 
		\includegraphics[width=1.0\linewidth]{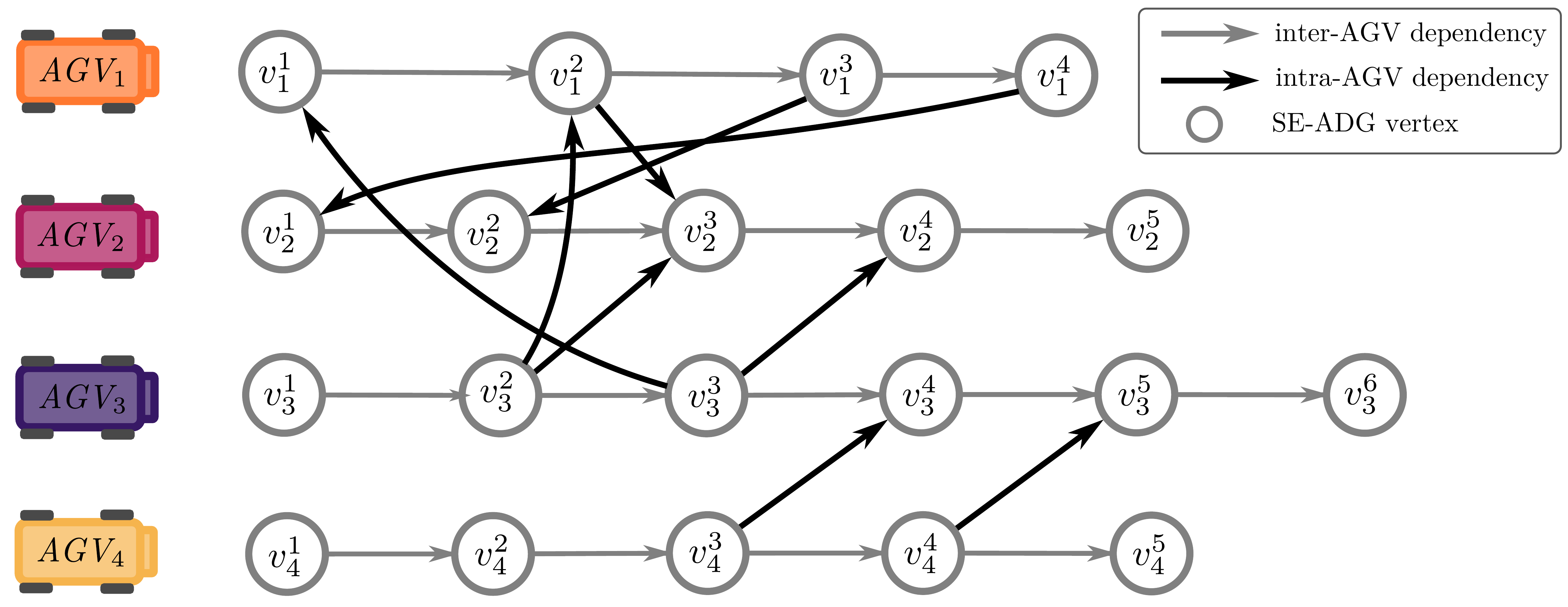}
		\caption{\acs{SE-ADG} $\SEADGgraph = (\SEADGvertices,\SEADGedges)$}
		\label{fig:adg}
	\end{subfigure}
	\caption{\textbf{Running example}: 
	  Roadmap $\mathcal{G} = (\mathcal{V},\mathcal{E})$ 
		occupied by $N=4$ \acs{AGVs} overlayed with the \acs{MAPF} plan $\mathcal{P}$
		indicated by the colored routes. 
	  \acs{AGV} ordering is indicated by the $z$-height relative to 
	    other routes, i.e., $\acs{AGV}_3$ before $\acs{AGV}_1$ before $\acs{AGV}_2$.
	  This ordering is more explicitly indicated by the
	  \acs{SE-ADG}, $\SEADGgraph = (\SEADGvertices,\SEADGedges)$, 
	  constructed 
	  from $\mathcal{P}$ using \algref{alg:seadg_algorithm}.
	}
	\label{fig:roadmap_and_adg}
\end{figure*}

\section{Coordinating Multiple \acs{AGV}s}
\label{sec:adg}

In this section, we introduce the concepts of a valid \acs{MAPF} plan as well as
  a formal introduction of the \acf{SE-ADG}, 
  a concept derived from the \acs{ADG} originally proposed 
  in \cite{hoenigPersistentRobustExecution2018}.
The \acs{SE-ADG} and properties introduced will form the foundation 
  of the methods introduced in subsequent sections.

\subsection{Valid \acs{MAPF} Plans}
Consider a workspace 
  represented by a graph $\mathcal{G} = (\mathcal{V},\mathcal{E})$
  which is occupied by a fleet of $N$ \acs{AGVs}, e.g., as in \figref{fig:roadmap}.
Each \acs{AGV} has a unique start and goal 
  position $s_i \in \mathcal{V}, s_i \neq s_j$ if $i \neq j \; \forall \; i,j \in \{1,\dots,N\}$ 
  and $g_i \in \mathcal{V}, g_i \neq g_j$ if $i \neq j \; \forall \; i,j \in \{1,\dots,N\}$, respectively. 
The task is for $\acs{AGV}_i$ to navigate from $s_i$ to $g_i$ 
  without collisions,
  $\forall \; i \; \in \{1,\dots,N\}$.
A solution to this task is called a 
  \acs{MAPF} solution which we represent as a set 
  $\mathcal{P} = \{\mathcal{P}_1,\dots,\mathcal{P}_N\}$,
  where $\mathcal{P}_i = \{ p_i^1, \dots, p_i^{N_i} \}$ 
  is a sequence of $N_i$ \textit{plan tuples} 
  representing the actions $\acs{AGV}_i$ must 
  take to navigate from $s_i$ to $g_i$. 
A \textit{plan tuple} $p_i^k = (\hat{v}(p_i^k), \hat{t}(p_i^k))$ 
  where the operators 
  $\hat{v}(p_i) : \mathcal{P}_i \to \mathcal{V}$,
  and $\hat{t}(p_i) : \mathcal{P}_i \to \mathbb{N}_0 $ 
  return the roadmap vertex and 
  \textit{planned} time when $\acs{AGV}_i$ 
  must be at vertex $\hat{v}(p_i)$,
  respectively.
Note that we consider the \textit{planned} time in a discrete fashion, 
  as used in almost all \acs{MAPF} formulations and 
  algorithms \cite{sternMultiAgentPathfindingDefinitions2019}.
\defref{def:mapf_solution} lists the conditions 
  for a valid \acs{MAPF} solution. 
Essentially, 
  for a \acs{MAPF} solution to be valid, 
  all \acs{AGVs} must reach
  their goals in finite time, and there can be  
  no collisions between the \acs{AGVs} along their planned routes.

\begin{mydefinition}[Valid \acs{MAPF} solution]
	A valid \acs{MAPF} solution is a set 
	  $\mathcal{P} = \{\mathcal{P}_1,\dots,\mathcal{P}_N\}$ 
	  such that the vertices $\hat{v}(p_i^k) \neq \hat{v}(p_j^l)$ 
	  if $\hat{t}(p_i^k) = \hat{t}(p_j^l)$, $k \in \{1,\dots,N_i\}$,
	  $l \in \{1,\dots,N_j\} \; \forall \; i,j \in \{1,\dots,N\}$. 
	Additionally, $\acs{AGV}_i$ and
	  $\acs{AGV}_j$ if $i \neq j$ must never traverse an edge 
	  $e \in \mathcal{E}$ in opposite directions in the same time-step.
	Finally, $\hat{v}(p_i^1) = s_i$ and $\hat{v}(p_i^{N_i}) = g_i \; \forall \; i,j \in \{1,\dots,N\}$.
	\label{def:mapf_solution}
\end{mydefinition}

\subsection{The \acl{SE-ADG}}

Although a valid \acs{MAPF} solution guarantees that all \acs{AGVs}
  reach their respective goals in finite time without collision, 
  the underlying assumption is that all \acs{AGVs} execute the
  plan without time-delays.
To relax this assumption, we introduce the \acf{SE-ADG},
  a graph-based data-structure used to define the ordering 
  of \acs{AGVs} as they navigate $\mathcal{G}$.
The idea is that a valid \acs{MAPF} solution can be used to generate 
  an \acs{SE-ADG} which can be used to execute the \acs{MAPF} plans 
  while maintaining collision-avoidance and route-completion guarantees.
  
Note that the \acs{SE-ADG} we present here is directly borrowed from 
  the original \acf{ADG} 
  presented in \cite{hoenigPersistentRobustExecution2018},
  but with the additional property that each vertex must involve a movement
  from two locations which are spatially exclusive of one another.

\begin{mydefinition}[\acl{SE-ADG}]
	An \acs{SE-ADG} is a directed graph 
	  $\SEADGgraph = (\SEADGvertices,\SEADGedges)$.
	$\SEADGvertices$ is a set of vertices
	  $v = (\{p_1,\dots,p_q\}, \textit{status})$ 
	  which define the movement of $\acs{AGV}_i$ ~from $\hat{v}(p_1)$, via 
	    intermediate locations, to $\hat{v}(p_q)$. 
	The variable $\textit{status} \in \{\staged, \inprogress, \completed \}$ indicates the current status of each movement.
	$\SEADGedges$ is a set of directed edges 
	  $e = (v, v')$ with $v, v' \in \SEADGvertices$.
	\label{def:adg}
\end{mydefinition}

The \acs{SE-ADG} represents the implicit ordering of the vertex visitation
  by each \acs{AGV} using a valid \acs{MAPF} plan $\mathcal{P}$ and 
  can be constructed using \algref{alg:seadg_algorithm}.
\mychange{Aside from the usage of different data structures,
  \algref{alg:seadg_algorithm} is practically identical to 
  the \acs{ADG} algorithm in \cite{hoenigPersistentRobustExecution2018},
  except for lines 7-13.} 
$loc(p_i) : \mathcal{P}_i \to \mathbb{R}^2$ returns location of plan tuple $p$,
  \mychange{$S_\text{AGV} \subset \mathbb{R}^2$ represents the area occupied by an \acs{AGV}}, 
  $\oplus$ is the Minkowski-sum.
Finally, $s(v) : \SEADGvertices \to \mathcal{V}$ returns the roadmap vertex 
  associated with the \textit{first} plan tuple in $v$; 
  $g(v) : \SEADGvertices \to \mathcal{V}$ the \textit{last} plan tuple in $v$.
$\hat{t}_g(v) : \SEADGvertices \to \mathbb{N}_0$ 
  returns $\hat{t}(p_q)$ where $p_q$ is the last plan tuple in $v$.
Once again, we use the hat to indicate that we are dealing with \textit{planned} times. 
The subscript $g$ in $\hat{t}_g(\cdot)$ is short for \textit{goal}.
\mychange{Going forward, we use the following notation to refer to \acs{AGVs}
  and \acs{SE-ADG} vertices:  $i$ and $j$ are both
 \acs{AGV} indices such that $i,j \in \{1,\dots, N\}$. 
  Furthermore, $k$ and $l$ are the \acs{SE-ADG} 
  vertex index of \acs{AGV} $i$ and $j$ respectively, i.e.,
  $k \in \{1,\dots,N_i\}$ and $l \in \{1,\dots,N_j\}$.}

\begin{algorithm}[b!]
	\caption{Compiling a \acf{SE-ADG}. Notable functional differences to the \acs{ADG} Algorithm in \cite{hoenigPersistentRobustExecution2018}
	in lines 7-13}
	\small
	\begin{algorithmic}[1]
		\renewcommand{\algorithmicrequire}{\textbf{Input:}}
		\renewcommand{\algorithmicensure}{\textbf{Result:}}
		\REQUIRE $\mathcal{P} = \{ \mathcal{P}_1,\dots,\mathcal{P}_N \}$ // valid \acs{MAPF} solution 
		\ENSURE  $\SEADGgraph$ \\ 
		\vspace{1mm}
		// Add sequence of events for each \acs{AGV}
		\FOR {$\mathcal{P}_i = \{ p_i^1, \dots, p_i^{N_i} \}$ \textbf{in} $\mathcal{P}$} 
		\STATE $p \leftarrow p_i^1$
		\STATE $v \leftarrow (\{p\}, \staged)$ 
		\STATE $v_\text{prev} \leftarrow \text{None}$
		\FOR {$k = 2$ to $N_i$}
		\STATE Append $p_i^k$ to sequence of plan tuples of $v$ \\
		// Check for spatial exclusivity
		\IF {$loc(p) \oplus S_\text{AGV} \cap loc(p_i^k) \oplus S_\text{AGV} = \emptyset$}
		\label{alg:adg_se}
		\STATE Add $v$ to $\SEADGvertices$
		\IF {$v_\text{prev}$ not \textit{None}}
		\STATE Add edge $e = (v_\text{prev}, v)$ to $\SEADGedges$
		\ENDIF
		\STATE $v_\text{prev} \leftarrow v$
		\STATE $p \leftarrow p_i^k$
		\STATE $v \leftarrow (\{p\}, \staged)$  \label{alg:adg_type_1_end}
		\ENDIF
		\ENDFOR
		\ENDFOR
		\vspace{1mm}
		// Add inter-AGV ordering constraints
		\FOR {$i = 1$ to $N$} \label{alg:adg_create_type_2}
		\FOR {$j = 1$ to $N$, $i \neq j$}
		\FOR {$k = 1$ to $N_i$}
		\FOR {$l = 1$ to $N_{j}$} \label{alg:adg_second_for_loop}
		\IF {$s(v_i^k) = g(v_j^l)$ \AND $\hat{t}_g(v_i^k) \leq \hat{t}_g(v_j^l)$} \label{alg:add_dep_2_1}
		\STATE Add edge $e = (v_i^k, v_j^l)$ to $\SEADGedges$ \label{alg:add_dep_2_2}
		\ENDIF
		\ENDFOR
		\ENDFOR
		\ENDFOR
		\ENDFOR
		\vspace{1mm}
		\RETURN $\SEADGgraph = (\SEADGvertices,\SEADGedges)$
	\end{algorithmic}
	\label{alg:seadg_algorithm}
\end{algorithm}

\algref{alg:seadg_algorithm} takes a valid \acs{MAPF} solution as input,
  and uses a two-stage approach to convert this into an \acs{SE-ADG}.
In the first stage (cf. lines 1-13), each \acs{AGV}'s plan is considered 
  individually. 
Spatial exclusivity of each vertex is guaranteed in line 7.  
The interaction between \acs{AGVs} is considered in the second stage
  (cf. lines 14-20).
Here, dependencies are generated between \acs{AGVs} if their planned routes 
  cover the same location at any point in time.

Initially, the \textit{status} 
  of $v_i^k$ is $\staged \; \forall \; i,k$.
The directed edges $e \in \SEADGedges$, 
  from here on referred to as dependencies, 
  define event-based constraints between two vertices.
Specifically, 
  $(v_i^k, v_{j}^{l})$ implies 
  that $v_{j}^{l}$ cannot be \inprogress{} 
  or \completed{} until $v_{i}^{k} = \completed$.
A dependency $(v_i^k, v_{j}^{l}) \in \SEADGedges$ 
  is classified as \textit{intra-AGV} if $i = j$ 
  and \textit{inter-AGV} if $i \neq j$.
If an \acs{SE-ADG} execution policy is 
  used to execute a valid \acs{MAPF} solution,
  an important property to ensure 
  finite-time task completion times for all \acs{AGVs}
  is that the \acs{SE-ADG} must be acyclic.
In this context, 
  finite-time task completion for all \acs{AGVs} is the same 
  as deadlock-free plan execution.

\mychange{The key difference between \algref{alg:seadg_algorithm} and 
  the \acs{ADG} algorithm in \cite{hoenigPersistentRobustExecution2018} 
  is the fact that 
  subsequent vertices are spatially exclusive cf. lines \ref{alg:adg_se}-\ref{alg:adg_type_1_end}.
The plans referenced in \cite{hoenigPersistentRobustExecution2018} 
  contain actions such as \emph{in-place rotations}.
Despite an \acs{AGV} not changing location by performing such an action,
  the \acs{ADG} will have two inter-\acs{AGV} edges.
Using spatial exclusivity, as in the \acs{SE-ADG}, 
  an \emph{in-place rotation} is merged with 
  a spatially transitional action in one vertex, 
  resulting in one inter-\acs{AGV} edge 
  to express this inter-\acs{AGV} dependency.}
Not only does this mean that the \acs{SE-ADG} has fewer vertices and edges,
  it will also be an important component for the  
  definition of a \textit{switched dependency} presented in the next section.

\subsection{\acs{SE-ADG} as a Plan Execution Policy}

In the following 
  we detail how an \acs{SE-ADG} can be used 
  as an execution policy to coordinate 
  the \acs{AGVs} and to accommodate for possible delays.
From \defref{def:adg}, we recall that the operator $\textrm{status}(v) : \SEADGvertices \to \textit{status}$ 
  returns the status of a vertex $v$. 
An \acs{AGV} is said to be \textit{executing} an \acs{SE-ADG} 
  vertex $v$ if it is performing the actions 
  defined by the plan tuple sequence $\{p_1,\dots,p_q\}$ of $v$.

\begin{mydefinition}[\acs{SE-ADG} plan execution policy]
	Consider a valid \acs{MAPF} plan $\mathcal{P}$ and 
	  the corresponding \acs{SE-ADG}, $\SEADGgraph$, 
	  constructed using \algref{alg:seadg_algorithm}.
	The \acs{SE-ADG}-execution policy is defined as follows:
	\begin{enumerate}
		\item Initially, $\text{status}(v) = \staged \; \; \forall v \in \SEADGvertices$;  
		\item Each $\acs{AGV}_i$'s first vertex is $v_i^1\; \forall \; i \; \in \; \{1,\dots,N\}$;
		\item $\acs{AGV}_i$ can only start \emph{executing} $v_i^k$ 
		if $\status{v} = \completed \; \forall \; e = (v,v_i^k) \in \SEADGedges$. 
	\end{enumerate}
	The \acs{SE-ADG} vertex statuses are updated by this policy:
	\begin{enumerate}
		\item $\status{v_i^k}$ changes from $\staged$ to $\inprogress$ 
		  if $\acs{AGV}_i$ is busy executing $v_i^k$;
		\item $\status{v_i^k}$ changes from $\inprogress$ to $\completed$
		  if $\acs{AGV}_i$ has finished executing $v_i^k$.
	\end{enumerate}
	\label{def:seadg_execution}
\end{mydefinition} 

\subsection{Properties of the \acs{SE-ADG}}

With the \acs{SE-ADG} execution policy from \defref{def:seadg_execution},
  we now show that if the \acs{SE-ADG} is acyclic, we can guarantee that 
  \acs{AGVs} can execute their plans in finite time in a collision-free manner.
In this context, we anticipate that 
  guaranteeing finite-time plan completion 
  is equivalent to ensuring 
  the plan execution is persistently deadlock-free.
First, we make the following assumption.

\begin{myassumption}
	A single \acs{AGV} can navigate the workspace 
	  (represented by roadmap $\mathcal{G}$) 
	  occupied by static and dynamic obstacles
	  in a collision-free manner using 
	  on-board navigation methods.
	  \label{ass:col_free_navigation}
\end{myassumption}

In the context of multiple \acs{AGVs} and \acs{MAPF}, 
  \assref{ass:col_free_navigation} is relatively nonconstraining.
It requires \acs{AGVs} to be able to follow the roadmap $\mathcal{G}$ and 
 navigate around or wait for static and dynamic third-party obstacles 
 which might partially or temporarily block its path, respectively.
This single \acs{AGV} navigation problem has already been addressed 
  in numerous works \cite{schoels2020nmpc, rosmann2017kinodynamic, williams2017model, triebel2016spencer}. 

If \assref{ass:col_free_navigation} is satisfied, guaranteeing collision-free 
  task execution in the multi-\acs{AGV} case requires us to ensure 
  that \acs{AGVs} do not collide with each other.
\mychange{Results in resource allocation 
  of concurrent system analysis such as \cite{Shrock2003},
  show that if a dependency graph is \textit{acyclic},
  plan execution is guaranteed to be deadlock-free.
In this context, if the \acs{SE-ADG} is
  constructed from a 
  valid \acs{MAPF} plan,
  we know that following the execution policy 
  in \defref{def:seadg_execution} will ensure each \acs{AGV} will complete its task 
  in a collision- and deadlock-free manner.}

\begin{mycorollary}[\acs{SE-ADG} guarantees collision-free plan execution]
Consider a valid \acs{MAPF} plan abstracted to 
  an \acs{SE-ADG} using \algref{alg:seadg_algorithm}.
\acs{AGVs} are guaranteed to execute the \acs{SE-ADG} plans
  in a collision-free manner if all \acs{AGVs} adhere to 
  the execution policy in \defref{def:seadg_execution}.
\label{res:collision_free_SEADG}
\end{mycorollary}
\begin{myproof}
\mychange{Consider the nominal execution of a \acs{MAPF} plan: 
    from \defref{def:mapf_solution},
	$\hat{v}(p_i^k) \neq \hat{v}(p_j^l)$ 
	if $\hat{t}(p_i^k) = \hat{t}(p_j^l)$,
	implying that no two \acs{AGVs} will occupy the same 
	location at the same time.
	Additionally, by \defref{def:mapf_solution},
	each $i$'th \acs{AGV} reaches its goal 
	$\hat{v}_g(p_i^{N_i}) \; \forall \; i \in \{1, \dots, N \}$.
Next, consider \algref{alg:seadg_algorithm} 
  lines \ref{alg:add_dep_2_1}-\ref{alg:add_dep_2_2} which ensure that 
  $\forall i,j \in 
  {1,\dots,N}$ and $k \in {1,\dots,N_i}$, $l \in {1,\dots,N_j}$ where 
  $s(v_i^k) = g(v_j^l)$ \textbf{and} $\hat{t}_g(v_i^k) \leq \hat{t}_g(v_j^l)$,
  a dependency $e = (v_i^k, v_j^l)$ is added to $\SEADGedges$.  
By the execution policy \defref{def:seadg_execution}, item 3,
  each \acs{AGV} will only move from a vertex $s(v)$ to $g(v)$ 
  if all edges pointing to $v$ have status completed. 
Because the \acs{AGVs} can only collide by visiting the same 
  location at the same time, and each instance of an \acs{AGV} 
  occupying the same location as another yields an edge in the 
  \acs{SE-ADG}, 
  no \acs{AGV} will occupy the same location at the same time,
  implying zero collitions during plan execution.}
\end{myproof}

\begin{mycorollary}[An acyclic \acs{SE-ADG} is sufficient to 
	guarantee deadlock- and collision-free plan execution]
	Consider an \acs{SE-ADG}, $\SEADGgraph$, 
	  constructed from a valid \acs{MAPF} plan $\mathcal{P}$ 
	  using \algref{alg:seadg_algorithm}.
	If $\SEADGgraph$ is 
	  acyclic and \assref{ass:col_free_navigation} holds, 
	each $\acs{AGV}_i$ will reach $g_{N_i}$ in finite time 
	  without collisions for all $i \in \{1,\dots,N\}$.
	\label{res:acyclic_SEADG_deadlock}
\end{mycorollary}
\begin{myproof} 
	Individually, the completion time of each vertex $v \in \SEADGvertices$ 
	  is finite by \assref{ass:col_free_navigation}.
	If $\SEADGgraph$ is acyclic, 
	  it has a topological ordering,
	implying that at each point, 
	  at least one \acs{SE-ADG} vertex can be executed,
	  until all vertices are \completed.
	This proves deadlock-free execution.
	Collision-free movement is proven in \corref{res:collision_free_SEADG}
\end{myproof}

Note that we are not able to extend \corref{res:acyclic_SEADG_deadlock} 
  to be a \textit{necessary} condition for deadlock-free plan execution, 
  since methods guaranteeing deadlock-free plan execution exist 
  which do not use an \acs{SE-ADG} approach.
Nevertheless, this \textit{sufficient} condition is still very useful, 
  as will be shown in subsequent sections.


Finally, we address the fact that not all 
  valid \acs{MAPF} solutions yield an acyclic \acs{SE-ADG}. 
A cyclic \acs{SE-ADG} comes from a plan which essentially 
  requires perfect synchronization among \acs{AGVs}.
However, with little limitation 
  to practical cases (see \remref{rem:acyclic_ADG_assumption}), we 
  assume that \acs{MAPF} solutions will yield acyclic \acs{SE-ADGs}.

\begin{myassumption}
  \acs{MAPF} problems are such that they initially yield an acyclic \acs{SE-ADG}.
\end{myassumption}

\begin{myremark}
Acyclicity of the \acs{SE-ADG} can be ensured 
	when the roadmap vertices outnumber the \acs{AGV} fleet size 
	i.e.$|\mathcal{V}| > N$ (as is typically the case in warehouse robotics)
	or a \acs{MAPF} solver such as kR-\acs{MAPF} is used with $k=1$\cite{atzmonRobustMultiAgentPath2020}. 
Alternatively, simple modification (e.g., an extra edge constraint in \acs{CBS}) 
	to existing \acs{MAPF} 
	solvers is sufficient to ensure acyclicity
	\cite{hoenigPersistentRobustExecution2018}.
\label{rem:acyclic_ADG_assumption}
\end{myremark}

%% file: sec/4_reordering.tex

\section{Reordering \acs{AGV} Plans: Introducing the \acl{SADG}}
\label{sec:sadg}

In \secref{sec:adg} we introduced the \acs{SE-ADG} and showed 
  that an acyclic \acs{SE-ADG} is a sufficient condition to guarantee 
  deadlock- and collision-free plan execution for multiple \acs{AGVs} 
  executing a valid \acs{MAPF} plan.
In this section, we address the core challenge we are tackling 
  throughout this manuscript:
  \textit{how can we adjust plans online to account for delays while maintaining 
  deadlock- and collision-free plan execution guarantees?}
To this end, 
  we introduce a new data-structure, 
  the \acf{SADG}, which facilitates the 
  systematic re-ordering of \acs{AGVs}
  based on time-delays.
We go on to show that the \acs{SADG} provides the ability 
  to maintain deadlock- and collision-free guarantees 
  of the original \acs{SE-ADG} on which it is based.

\subsection{Switched Dependencies}

Before introducing the \acs{SADG}, we need to introduce 
  the fundamental building block on which it is based: the \textit{switched dependency}.
Consider an inter-\acs{AGV} (i.e. $i \neq j$) dependency 
  $e_\text{original} = (v_j^l,v_i^k) \in \SEADGedges$.
From here on, we refer to an \textit{inter-\acs{AGV} dependency} 
  simply as a \textit{dependency}.
  As per \defref{def:seadg_execution}, $e_\text{original}$ 
  implies $\text{status}(v_j^l) = \textit{completed}$ 
  before $\text{status}(v_i^k) = \textit{in-progress}$ when executing the \acs{SE-ADG}-based plans.
In terms of the roadmap $\mathcal{G}$, this is equivalent to requiring 
  $\acs{AGV}_j$ to leave $s(v_j^l)$ (reach $g(v_j^l)$) before $\acs{AGV}_i$ can advance to $g(v_i^k)$.
Note the implicit ordering that $\acs{AGV}_j$ must go to $s(v_j^l) = g(v_i^k)$ before $\acs{AGV}_i$.
The idea of a switched dependency is to reverse this implicit ordering 
  while ensuring \acs{AGVs} don't occupy the same vertex in $\mathcal{G}$ at the same time.

\begin{figure}[t!]
	\centering
	\begin{tikzpicture}
	\begin{scope}[every node/.style={draw=gray, fill=colorinside, thick, ellipse, minimum size=1.5pt}]
	\node (Ai) at (0.0, 1.4) {$v_i^{k-1}$};
	\node (Bi) at (2.4, 1.4) {$v_i^{k}$};
	\node (Ci) at (4.0, 1.4) {$v_i^{k+1}$};
	\node (Aj) at (0.0, 0.0) {$v_j^{l-1}$};
	\node (Bj) at (1.6, 0.0) {$v_j^{l}$};
	\node (Cj) at (4.0, 0.0) {$v_j^{l+1}$};
	\end{scope}
	\begin{scope}[every node/.style={fill=none,thin,draw=none}]
	\node (Li) at (-1.5, 1.4) {};
	\node (Ri) at (5.5, 1.4) {};
	\node (Lj) at (-1.5, 0.0) {};
	\node (Rj) at (5.5, 0.0) {};
	\node (Legend1L) at (-0.5, -1.0) {};
	\node (Legend1R) at (0.5, -1.0) {};
	\node (Legend2L) at (2.0, -1.0) {};
	\node (Legend2R) at (3.0, -1.0) {};
	\node (Legend1) at (1.2, -1.0) {\textit{forward}};
	\node (Legend2) at (3.7, -1.0) {\textit{reverse}};
	\end{scope}
	\begin{scope}[>={Stealth[lightgray]},
	every node/.style={fill=none,circle},
	every edge/.style={draw=lightgray,very thick}]
	\path [->] (Ai) edge node {} (Bi);
	\path [->] (Bi) edge node {} (Ci);
	\path [->] (Aj) edge node {} (Bj);
	\path [->] (Bj) edge node {} (Cj);
	\end{scope}
	\begin{scope}[>={Stealth[lightgray]},
	every node/.style={fill=none,circle},
	every edge/.style={draw=lightgray,very thick, path fading=east}]
	\path [-] (Ci) edge node {} (Ri);
	\path [-] (Cj) edge node {} (Rj);
	\end{scope}
	\begin{scope}[>={Stealth[lightgray]},
	every edge/.style={draw=lightgray,very thick, path fading=west}]
	\path [->] (Li) edge node {} (Ai);
	\path [->] (Lj) edge node {} (Aj);
	\end{scope}
	\begin{scope}[>={Stealth[purple]},
	every node/.style={fill=none,circle},
	every edge/.style={draw=purple,very thick}]
	\path [->] (Cj) edge node {} (Ai);
	\path [->] (Legend2L) edge node {} (Legend2R);
	\end{scope}
	\begin{scope}[>={Stealth[black]},
	every node/.style={fill=none,circle},
	every edge/.style={draw=black,very thick}]
	\path [->] (Bi) edge node {} (Bj);
	\path [->] (Legend1L) edge node {} (Legend1R);
	\end{scope}
	\end{tikzpicture}
	\caption{Visual illustration of a switched dependency: 
	Original dependency (black) and its reversed counterpart (red).}
	\label{fig:reverse_dependency}
\end{figure}
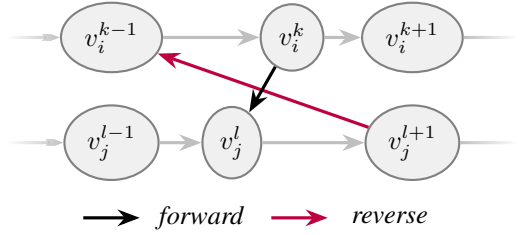

\begin{mydefinition}[Switched inter-\acs{AGV} dependency]
	Given a dependency $e_\text{fwd} = (v_i^k, v_j^l) \in \SEADGedges$,
	a switched dependency is an edge $e_\text{rev} = (v,v')$ which 
	   ensures $\acs{AGV}_i$ reaches $s(v_j^l) = g(v_i^k)$ before $\acs{AGV}_j$ without collision.
	\label{def:reverse_dependency}
\end{mydefinition}

Given a dependency $e_\text{fwd}$, a switched dependency which fulfills 
  \defref{def:reverse_dependency} can be determined using \lemref{lem:reverse_dependency}.
A dependency and its reversed counterpart 
  are illustrated in \figref{fig:reverse_dependency}.

\begin{mylemma}[Switched inter-\acs{AGV} dependency]
	Let 
	  $v^k_i, v^l_j$, $v^{l+1}_{j}, v^{k-1}_{i} \in \SEADGvertices$,
	  where $\SEADGgraph = (\SEADGvertices,\SEADGedges)$
	  from \defref{def:adg}.
	Then $d' = (v^{l+1}_{j}, v^{k-1}_{i})$ is the 
	  switched counterpart of $d = (v^k_i,v^l_j)$ which adheres to 
	  \defref{def:reverse_dependency}.
	  \label{lem:reverse_dependency}
\end{mylemma}
\begin{myproof}
	The dependency $d = (v^k_i,v^l_j)$ 
	  encodes the constraint $t_s(v_{j}^{l}) \geq t_g(v_i^k)$.
	The switched counterpart of $d$ 
	  is denoted as $d' = (v^{l+1}_{j}, v^{k-1}_{i})$.
	  $d'$ encodes 
	  the constraint $t_s(v_{i}^{k-1}) \geq t_g(v_j^{l+1})$.
	By definition, 
	  $t_s(v_i^k) \geq t_g(v_i^{k-1}) $ 
	  and $ t_s(v_j^{l+1}) \geq t_g(v_j^l) $.
	Since $t_g(v) \geq t_s(v)$, 
	  this implies that $d'$ 
	  encodes the constraint 
	  $ t_s(v_i^k) \geq t_g(v_{j}^{l}) $, 
	  satisfying \defref{def:reverse_dependency}. 
\end{myproof}

Note that, as discussed in \secref{sec:adg}, 
  the \acs{SE-ADG} is spatially exclusive, allowing us to use only a single
  dependency as the switched counterpart of one dependency.

\subsection{The \acl{SADG}}

\lemref{lem:reverse_dependency} provides us with 
  a method to maintain 
  collision avoidance while 
  re-ordering \acs{AGVs}.
We are now in the position to extend the \acs{SE-ADG} 
  to enable the re-ordering of \acs{AGVs} 
  using \textit{switchable dependencies}.
To this end, we introduce the \acf{SADG}, 
  a mapping from 
  a binary vector, \textit{\textbf{b}}, to an \acs{SE-ADG}, $\SEADGgraph$.

\begin{mydefinition}[\acl{SADG}]
	An \acs{SADG} 
	  is a mapping 
	$\SADGgraph{\bvec} : \{0,1\}^{m_T} \to \SEADGgraphset$ 
	which outputs the resultant \acs{SE-ADG} 
	based on the dependency selection 
	represented by $\bvec = \{b_1,\dots,b_{m_T}\}$,
	where $b_m = 0$ and $b_m = 1$ imply 
	selecting the forward and reverse 
	dependency of pair $m$, respectively, $m \in \{1,\dots,m_T\}$.
	\label{def:sadg}  	
\end{mydefinition}

In \defref{def:sadg}, $\SEADGgraphset$ refers 
  to the set of all possible $\SEADGgraph$'s.
Depending on the value of $\bvec$, $\SADGgraph{\bvec}$ will result in a different 
  $\SEADGgraph$.
Similarly to the \acs{SE-ADG}, 
  an \acs{SADG} can be constructed from a
  valid \acs{MAPF} solution $\mathcal{P}$
  using \algref{alg:sadg_algorithm}.
Like \algref{alg:seadg_algorithm}, \algref{alg:sadg_algorithm} also consists of 
  two-stages.
\mychange{In fact, the only difference between \algref{alg:sadg_algorithm}
  and \algref{alg:seadg_algorithm}
  is in lines 19-25.
Instead of just creating the dependency $\edgeFWD$,
  a check is done to validate 
  if its reverse counterpart,
  $\edgeREV$, 
  can be constructed (cf. line \ref{alg:sadg_2})}.
If so, a binary switching function  
  $\binaryMapFunc : \{0,1\} \to \SEADGedges$ 
  is appended to the set of dependencies $\SADGedges{\bvec}$ 
  (cf. line \ref{alg:sadg_1}), where,
  as per \defref{def:sadg}, $\binaryMap{b_m=0}{\edgeFWD}{\edgeREV} = \edgeFWD$,
  and $\binaryMap{b_m=1}{\edgeFWD}{\edgeREV} = \edgeREV$.
Conversely, if the reverse counterpart of $\edgeFWD$ cannot be constructed,
  dependency switching is not possible for $\edgeFWD$ and 
  there is no need for a binary variable $b_m$.
  In this case, the dependency $\edgeFWD$ is appended to $\SADGedges{\bvec}$ (cf. line \ref{alg:sadg_3}).

\begin{figure}[h!]
	\centering
	\includegraphics[width=1.0\linewidth]{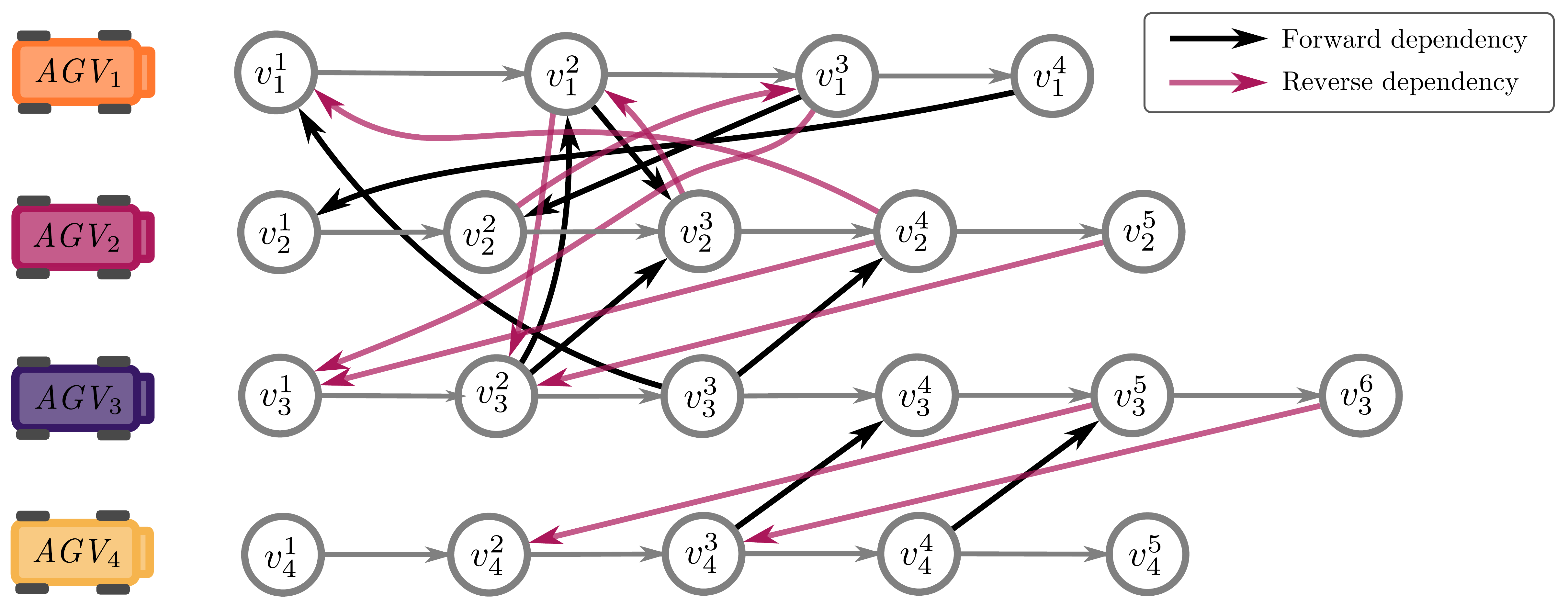}
	\caption{The \acs{SADG} $\SADGgraph{\bvec}$ constructed from the same valid \acs{MAPF} 
		solution used to construct the \acs{SE-ADG} in \figref{fig:adg}.
		}
	\label{fig:sadg}
\end{figure}

\begin{algorithm}[t!]
	\caption{Compiling a \acl{SADG}}
	\small
	\begin{algorithmic}[1]
		\renewcommand{\algorithmicrequire}{\textbf{Input:}}
		\renewcommand{\algorithmicensure}{\textbf{Result:}}
		\REQUIRE $\mathcal{P} = \{ \mathcal{P}_1,\dots,\mathcal{P}_N \}$ // valid \acs{MAPF} solution
		\ENSURE  $\SADGgraph{\bvec}$ \\
		\vspace{2mm}
		// Add sequence of events for each \acs{AGV}
		\FOR {$\mathcal{P}_i = \{ p_i^1, \dots, p_i^{N_i} \}$ \textbf{in} $\mathcal{P}$}   \label{alg:sadg_type_1_begin}
		\STATE $p \leftarrow p_i^1$
		\STATE $v \leftarrow (\{p\}, \textit{staged})$ 
		\STATE $v_\text{prev} \leftarrow$ \textit{None}
		\FOR {$k = 2$ to $N_i$}
		\STATE Append $p_i^k$ to sequence of plan tuples of $v$ \\
		// Check for spatial exclusivity
		\IF {$loc(p) \oplus S_\text{AGV} \cap loc(p_i^k) \oplus S_\text{AGV} = \emptyset$}
		\label{alg:adg_construction_spatial_exclusivity}
		\STATE Add $v$ to $\SADGvertices{\bvec}$
		\IF {$v_\text{prev}$ not \textit{None}}
		\STATE Add $(v_\text{prev}, v)$ to $\SADGedges{\bvec}$
		\ENDIF
		\STATE $v_\text{prev} \leftarrow v$
		\STATE $p \leftarrow p_i^k$
		\STATE $v \leftarrow (\{p\}, staged)$  \label{alg:sadg_type_1_end}
		\ENDIF
		\ENDFOR 
		\ENDFOR
		\vspace{2mm}
		// Add switchable dependency pairs
		\FOR {$i = 1$ to $N$} 
		\label{alg:sadg_create_type_2}
		\FOR {$j = 1$ to $N$, $j \neq i$}
		\FOR {$k = 1$ to $N_i$}
		\FOR {$l = 1$ to $N_{j}$} \label{alg:sadg_second_for_loop}
			\IF {$s(v_i^k) = g(v_j^l)$ \AND $\hat{t}_g(v_i^k) \leq \hat{t}_g(v_j^l)$}
				\STATE $\edgeFWD \leftarrow (v_i^k, v_j^l)$
				\IF {\big( $v_i^{k-1}$, $v_j^{l+1} \big) \in \SADGvertices{\bvec}$} \label{alg:sadg_2}
				\STATE $\edgeREV \leftarrow (v_j^{l+1}, v_i^{k-1})$
				\STATE Add $\binaryMap{b_m}{\edgeFWD}{\edgeREV}$ to $\SADGedges{\bvec}$ \label{alg:sadg_1}
				\ELSE
				\STATE Add $\edgeFWD$ to $\SADGedges{\bvec}$ \label{alg:sadg_3}
				\ENDIF
			\ENDIF
		\ENDFOR
		\ENDFOR
		\ENDFOR
		\ENDFOR
		\vspace{1mm}
		\RETURN $\SADGgraph{\bvec} = \big( \SADGvertices{\bvec}, \SADGedges{\bvec} \big)$
	\end{algorithmic} 
	\label{alg:sadg_algorithm}
\end{algorithm}
\figref{fig:sadg} shows an illustration of the \acs{SADG} for the running example
  illustrated in \figref{fig:roadmap_and_adg}.
Note that all the forward dependencies (solid, black arrows), are identical 
  to those in \figref{fig:adg}.
This is because \algref{alg:sadg_algorithm} follows the same process to generate 
  the forward dependencies as \algref{alg:seadg_algorithm}.
Note also how the forward dependency $\edgeFWD = (v_1^3, v_3^1)$ has no reverse counterpart.
This is because its reverse counterpart, 
  as specified by \lemref{lem:reverse_dependency},
  would be $\edgeREV = (v_1^4, v_3^0)$, which does not exist 
  (as checked in \algref{alg:sadg_algorithm}, line \ref{alg:sadg_2}).

\subsection{\acs{SADG} Properties and Execution Policy}

With the aim of introducing an \acs{SADG}-based control 
  scheme in \secref{sec:shc}, we need to ensure the 
  \acs{SE-ADG} execution policy in \defref{def:seadg_execution}
  can be applied to \acs{SADGs}.
To this end, recall that $\SADGgraph{\bvec}$ yields an \acs{SE-ADG} for a particular
  $\bvec$.
When all \acs{AGVs} are at their starting positions
  and $\bvec = \zerovec$,
  $\SADGgraph{\bvec} = \SEADGgraph$,
  the same \acs{SE-ADG} as obtained with \algref{alg:seadg_algorithm},
  which was shown to ensure collision-free and deadlock-free 
  plans in \corref{res:acyclic_SEADG_deadlock}.

Although we have proven collision- and deadlock-avoidance 
  if \acs{AGVs} follow the execution 
  policy in \defref{def:seadg_execution}, 
  this only holds if the underlying \acs{SE-ADG} remains constant.
We will now show that if $\bvec$ is chosen such that the
  all edges in $\SADGedges{\bvec}$ in 
  $\SEADGgraph = \SADGgraph{\bvec}$ do not violate the 
  assumptions in the execution policy of \defref{def:seadg_execution},
  \corref{res:collision_free_SEADG} can be extended to \acs{SADGs}.

\begin{mycorollary}[Persistent collision-free plan execution for \acs{SADGs}]
  For varying $\bvec$, the resultant \acs{SE-ADG} from $\SADGgraph{\bvec}$  
    will guarantee collision-free plan execution
  as long as
  \begin{enumerate}
    \item $\bvec$ is chosen such that $\forall \; e = (v_i^k,v_j^l) \; \in \SADGedges{\bvec}$
    the head of $e$, $v_j^l$, has $\status{v_j^l} = \staged$ at the time $\bvec$ is changed.
    \item \acs{AGVs} follow the plan execution 
      policy \defref{def:seadg_execution} based on the changing $\SADGedges{\bvec}$ at all times.
  \end{enumerate} 
  \label{res:persistent_collision_free_SADG}
\end{mycorollary}
\begin{myproof}
  Proof by induction.
  Initially, at time $t = T_0$, $\bvec = \zerovec$ and $\SEADGgraph$ is acyclic 
    which guarantees collision-free 
    plan execution by \corref{res:collision_free_SEADG}.
  Consider at time $t = T_1 > T_0$, $\bvec$.
  Let $\bvec_\text{switched} \subset \bvec$ refer to the subset of 
    $\bvec$ that is different between $t \leq T_1$ and $t > T_1$.
  Since $\forall \; b \in \bvec_\text{switched}$,
    $\status{v_j^l} = \staged$, $\status{v_j^{l'}} \; \forall l' \geq l$,
    by the construction of intra-\acs{AGV} dependencies in \algref{alg:seadg_algorithm},
      cf. lines 1-13. 
  Hence, at time $t>T_1$, the constraints imposed by all the 
    newly active edges introduced by $\bvec_\text{switched}$ 
    have not been violated.
  Since all constraints are adhered, collision-avoidance is guaranteed.
  The same logic applies for a subsequent switching at time $t = T_2 > T_1$,
    thus proving persistent collision-avoidance.
\end{myproof}

Note that although \corref{res:persistent_collision_free_SADG} 
  provides us constraints on the changing of $\bvec$ 
  at any time to ensure persistent collision-avoidance guarantees,
  no guarantees are made regarding deadlocks.
Hence, it is possible that, following \corref{res:persistent_collision_free_SADG},
  $\bvec$ is chosen which causes the \acs{AGVs} to enter into a deadlock.
Viewed from the perspective of the \acs{SE-ADG}: 
  it is possible that a value of $\bvec$
  causes a cycle in the resultant \acs{SE-ADG}. 
Finding a value of $\bvec$ which also guarantees deadlock-free 
  plans, i.e. persistent acyclicity of $\SADGgraph{\bvec}$,
  will be considered in \secref{sec:shc}. 

\begin{figure}[h!]
	\centering
	\includegraphics[width=1.0\linewidth]{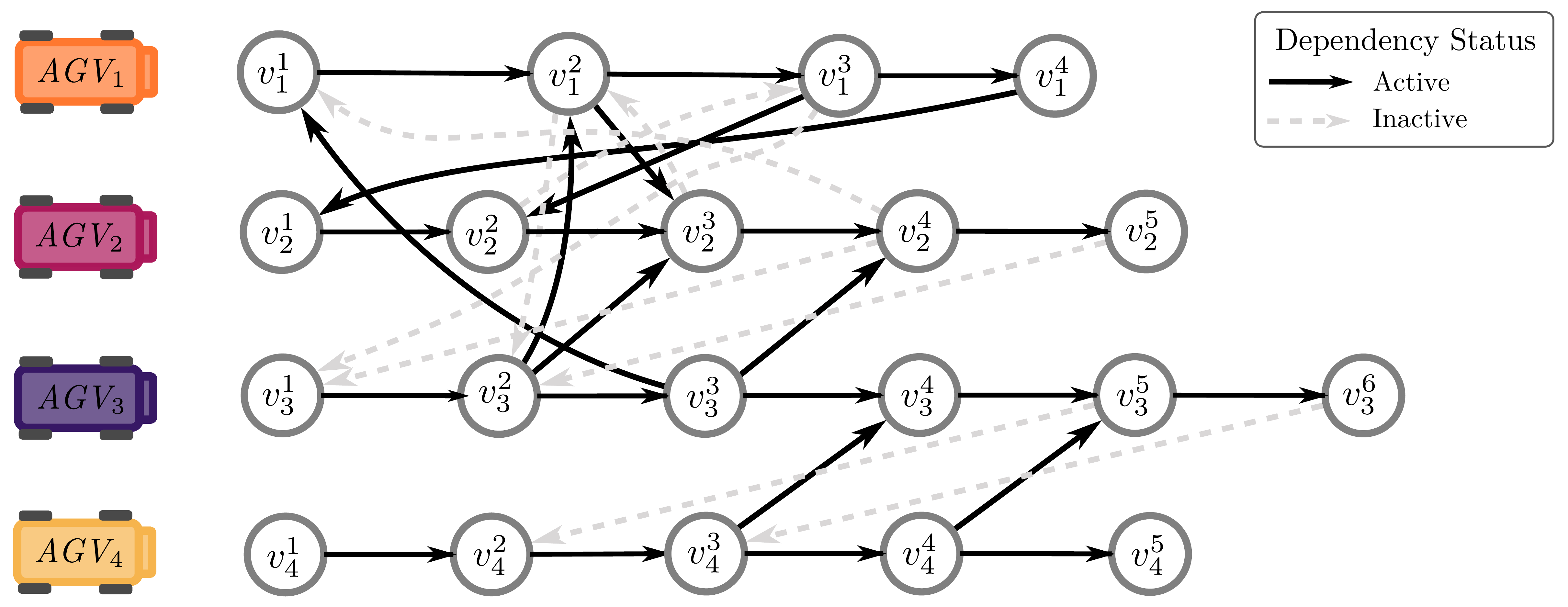}
	\caption{Illustration of $\SADGgraph{\bvec}$ for $\bvec = \zerovec$. 
	    The resultant $\SEADGgraph$ is depicted with solid black arrows.
		The dependencies not part of $\SADGgraph{\zerovec}$ are indicated 
		by dotted lines.}
	\label{fig:active_seadg}
\end{figure}

For our running example, 
  in the case that $\bvec = \zerovec$,
  the active \acs{SE-ADG} 
  is shown in \figref{fig:active_seadg}.
Note that \figref{fig:active_seadg} is practically identical to 
  \figref{fig:adg} because $\bvec = \zerovec$
  corresponds to the original \acs{SE-ADG} constructed using \algref{alg:seadg_algorithm}.
  
%% file: sec/5_ocp_feedback.tex

\section{\acl{SHC} Scheme}
\label{sec:shc}

\begin{figure*}[t!]
	\centering
	\includegraphics[width=0.95\textwidth]{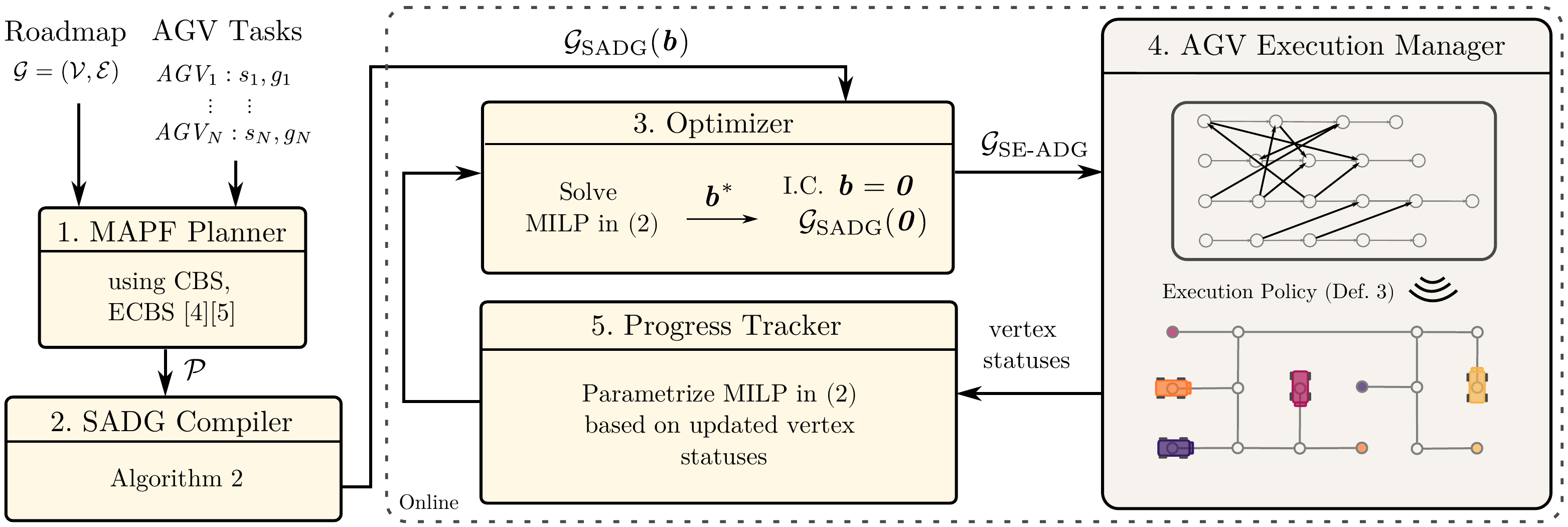}
	\vspace{2mm}
	\caption{\textbf{Shrinking horizon feedback control diagram:} 
	The planning 
	  phase consists of \textit{1. \acs{MAPF} Planner} and \textit{SADG Compiler} which
	  yields an \acs{SADG} given a roadmap and \acs{AGV} tasks.
	  Initially, in \textit{3. Optimizer}, the \acs{SADG} is initialized 
	    with $\bvec = \zerovec$ yielding an initial \acs{SE-ADG}
	  which can be used by \acs{AGVs} to execute their plans. 
	  As \acs{AGVs} progress,
	    and are inevitably delayed, their status 
		is tracked in \textit{5. Progress Tracker}, 
	  and an optimization 
		iteration is performed to re-order the \acs{AGVs} 
		in \textit{3. Optimizer}.
	  The solution to the \acs{MILP} is used to update the active \acs{SE-ADG}
		used for plan execution in a feedback loop. 
	  The feedback loop runs at a sampling frequency of $h$.}
	\label{fig:shc_strategy_diagram}
	\end{figure*}

Consider \acs{AGVs} executing their respective plans,
  as described by an \acs{SE-ADG} $\SEADGgraph^0 = \SADGgraph{\zerovec}$,
  adhering to execution policy in \defref{def:seadg_execution}.
In the case that any subset of the \acs{AGVs} is delayed,
  the cumulative route completion times of the \acs{AGVs} 
  could be reduced if the \acs{SE-ADG} is modified at a time $T_1 > T_0$,
  where $T_0$ refers to the time the \acs{AGVs} started executing the plans.
In \secref{sec:sadg}, we showed that 
  if, at time $T_1$, $\bvec$ is chosen in accordance with 
  \corref{res:persistent_collision_free_SADG}, 
  persistent collision-avoidance of the \acs{AGVs} is guaranteed 
  when following the 
  execution policy in \defref{def:seadg_execution}.
However, \corref{res:persistent_collision_free_SADG} 
  does not guarantee plan completion (i.e. deadlock-free movement).
This is because changing $\bvec$ at $T_1$
  could result in a cyclic \acs{SE-ADG}, causing a deadlock.
Therefore, the objective is to find $\bvec^*$ at $T_1$ to ensure
  $\SEADGgraph^* = \SADGgraph{\bvec^*}$ is acyclic,
  while minimizing the route-completion time of the \acs{AGV} fleet,
  based on the individual \acs{AGV} delays at time $T_1$.
  
In this section, we show that finding $\bvec^*$ is equivalent to 
  solving an \acf{OCP}.
We go on to solve this \acs{OCP} using a \acf{MILP} formulation,
  which we integrate into a shrinking horizon feedback control scheme.

\subsection{\acl{OCP}}

At any time during the execution of their respective plans,
  given an initial \acs{SE-ADG}, $\SADGgraph{\bvecstart}$,
  the \acs{OCP} can be formulated as follows
\begin{subequations}
	\begin{align}
		& \optOCP{\bvec} = \min_{\bvec, \tgvec, \tsvec} \sum_{i=1}^{N} t_g(v_i^{N_i}) \label{eq:ocp_shc_cost_fun} \\
		  						& \hspace{5mm} \text{s.t.} \nonumber \\ 
		t_g(v_i^k) &> t_s(v_i^k) + \Delta T(v_i^k) \; \forall \; v_i^k \in \Pi \big( \SADGvertices{\bvec} \big), \label{eq:ocp_shc_1_1} \\ 
		t_s(v_i^{k+1}) &> t_g(v_i^k)   \; \forall \; v_i^{k+1} \in \Gamma \big( \SADGvertices{\bvec} \big), \label{eq:ocp_shc_1_2} \\
		\status{v'} &= \staged \;\; \forall \; (v,v') \; \in \Psi(\bvec, \bvecstart), \label{eq:ocp_shc_2_1} \\
		t_g(v_i^k) &< t_s(v_j^l) \;\; \forall \; (v_i^k, v_j^l) \in \SADGedges{\bvec} \; \text{if} \;\; i \neq j,  \label{eq:ocp_shc_2_2} 
	\end{align}
\label{eq:ocp_shc}
\end{subequations}
where $\Psi\big(\bvec, \bvecstart \big)$ returns only the edges 
  in $\SADGedges{\bvec}$ which do not exist in $\SADGedges{\bvecstart}$,
\begin{align*}
	\Psi\big(\bvec, \bvecstart \big) = \{ e \; | \; e \in \SADGedges{\bvec} \land e \notin \SADGedges{\bvecstart} \}.
\end{align*}
Furthermore, $\Pi(\cdot)$ and $\Gamma(\cdot)$ are filters such that
\begin{align*}
	\Pi\big(\mathcal{V}\big) &= \big\{ v \in \mathcal{V} \; | \; \status{v} \in \{ \staged, \inprogress \} \big\}, \\
	\Gamma\big(\mathcal{V}\big) &= \big\{ v \in \mathcal{V} \; | \; \status{v} = \staged \big\}.
\end{align*}
Finally, $\Delta T(v_i^k)$ is the estimated time $\acs{AGV}_i$ will
  take to complete $v_i^k$, defined as 
\begin{align*}
	\Delta T(v_i^k) = 
	\begin{cases}
		T_\text{est}(v_i^k) &\text{ if } \status{v_i^k} = \staged, \\
		\mu T_\text{est}(v_i^k) &\text{ if } \status{v_i^k} = \inprogress, \\
		0 &\text{ if } \status{v_i^k} = \completed, \\
	\end{cases}
\end{align*}
where $T_\text{est}(v_i^k)$ is the \textit{total} 
  estimated time it will take 
  $\acs{AGV}_i$ to complete $v_i^k$,
  and $\mu \in [0,1]$ is the fraction of $v_i^k$ 
  that still needs to be completed.
Since $t_g(v_i^{N_i})$ refers to the time where 
$\acs{AGV}_i$ will reach its goal position,
  the cost function in \eqnref{eq:ocp_shc_cost_fun} 
  is the cumulative route completion time of all $N$ \acs{AGVs}.
Note that \eqnref{eq:ocp_shc_1_1} and \eqnref{eq:ocp_shc_1_2} 
  enforce the route sequence of each individual \acs{AGV},
  whereas \eqnref{eq:ocp_shc_2_1} and \eqnref{eq:ocp_shc_2_2} 
  enforce ordering constraints between \acs{AGVs}.
Moreover, \eqnref{eq:ocp_shc_2_1}
  ensures the heads of all \textit{switched} dependencies 
  point to \staged{} vertices. 

\mychange{
\begin{mycorollary}[Cyclic \acs{SE-ADG} yields constraint violation]
A cycle in the \acs{SE-ADG} violates constraints of 
  \acs{OCP} \eqnref{eq:ocp_shc} and therefore any feasible solution of \acs{OCP} \eqnref{eq:ocp_shc}
  is acyclic.
	\label{res:cyclic_equals_infeasible}
\end{mycorollary}
\begin{myproof}
Without loss of generality, consider the cyclic dependency chain 
  formed by a dependency from $v_i^k$ to $v_j^l$, 
  and from $v_j^l$ back to $v_i^k$.
The dependencies forming this cycle translate to 
  the following constraints, as specified in \eqnref{eq:ocp_shc_2_2},
\begin{subequations}
	\begin{align}
	v_i^k & \to v_j^l \; : \; t_g(v_i^k) < t_s(v_j^l), \\
		v_j^l & \to v_i^k \; : \; t_g(v_j^l) < t_s(v_i^k),
	\end{align}
\label{eq:cyclicity_proof_1}
\end{subequations}
Furthermore, by \eqnref{eq:ocp_shc_1_1}, we require that
\begin{subequations}
	\begin{align}
		t_g(v_i^k) & > t_s(v_i^k) + \Delta T(v_i^k), \\
		t_g(v_j^l) & > t_s(v_j^l) + \Delta T(v_j^l),
	\end{align}
\label{eq:cyclicity_proof_2}
\end{subequations}
Since $\Delta T(v_*^*) \geq 0$, 
  observe that \eqnref{eq:cyclicity_proof_1} and \eqnref{eq:cyclicity_proof_2} 
  lead to the contradiction that both $t_g(v_i^k) < t_s(v_l^j)$ and
  $t_g(v_i^k) > t_s(v_l^j)$ must hold.
Such a contradiction appears for every (possibly longer) cycle within the \acs{SE-ADG}.
This result directly implies that any feasible solution to \acs{OCP} \eqnref{eq:ocp_shc}
  is acyclic.
\end{myproof}

Next, we show that if the initial \acs{SE-ADG} is acyclic,
  the \acs{OCP} is feasible and in turn yields an acyclic \acs{SE-ADG}.
}
\begin{mycorollary}[A solution to \eqnref{eq:ocp_shc} 
	exists if $\SADGgraph{\bvecstart}$ is acyclic]
	If $\SADGgraph{\bvecstart}$ is acyclic, 
	  the minimizer to \eqnref{eq:ocp_shc}, $\bvec^*$, 
	  exists, $\optOCP{\bvec^*}$ is finite.,
	  implying $\SADGgraph{\bvec^*}$ is acyclic.
	\label{res:acyclic_seadg_opt_has_solution}
\end{mycorollary}
\begin{myproof}
	A direct result from \corref{res:acyclic_SEADG_deadlock} is that
	  if $\SADGgraph{\bvecstart}$ is acyclic,
	  the route completion time of all \acs{AGVs} is finite.
	Because the cost function of \eqnref{eq:ocp_shc} equals the cumulative 
	  route completion time of all \acs{AGVs}, 
	  $\optOCP{\bvecstart}$ is necessarily finite and $\bvecstart$ 
	  is a solution to \eqnref{eq:ocp_shc}.
	  \mychange{Consequently, the minimizer $\bvec^*$ exists 
	  and is a solution of \eqnref{eq:ocp_shc}.
	From \corref{res:cyclic_equals_infeasible},
	  this means that $\SADGgraph{\bvec^*}$ is acyclic.
	  }
\end{myproof}

\subsection{Formulation as \acl{MILP}}
Working towards the definition of a feedback control 
  scheme, we now formulate the \acs{OCP} in \eqnref{eq:ocp_shc}
  as an \acs{MILP} as follows
\begin{subequations}
	\begin{align}
		& \min_{\bvec, \tgvec, \tsvec} \sum_{i=1}^{N} t_g(v_i^{N_i}) \label{eq:milp_shc_cost_fun} \\
		  						& \hspace{5mm} \text{s.t.} \nonumber \\ 
		t_g(v_i^k) &> t_s(v_i^k) + \Delta T(v_i^k) \; \forall \; v_i^k \in \Pi \big( \SADGvertices{\bvec} \big), \label{eq:milp_shc_1_1} \\ 
		t_s(v_i^{k+1}) &> t_g(v_i^k)   \; \forall \; v_i^{k+1} \in \Gamma \big( \SADGvertices{\bvec} \big), \label{eq:milp_shc_1_2} \\
		\status{v'} &= \staged \;\; \forall \; (v,v') \; \in \Psi(\bvec, \bvecstart), \label{eq:milp_shc_2_1} \\
		t_g(v_i^k) &< t_s(v_j^l) \;\; \forall \; (v_i^k, v_j^l) \in \SADGedges{\bvec} \; \text{if} \;\; i \neq j,  \label{eq:milp_shc_2_2} \\
		t_s(v_j^l) &> t_g(v_i^k) - b M, \; \forall \; \; b \in \bvec, \label{eq:milp_shc_big_M_1} \\
		t_s(v_i^{k-1}) &> t_g(v_{j}^{l+1}) - (1 - b) M, \; \forall \; \; b \in \bvec, \label{eq:milp_shc_big_M_2}
	\end{align}
\label{eq:milp_shc}
\end{subequations}
where $M \gg 0$ is a large constant and
\mychange{$i$ and $j$ are both
 indices referring to a specific \acs{AGV} 
 such that $i,j \in \{1,\dots, N\}$. 
 Furthermore, $k$ and $l$ are the \acs{SE-ADG} 
 vertex index of \acs{AGV} $i$ and $j$ respectively, i.e.,
 $k \in \{1,\dots,N_i\}$ and $l \in \{1,\dots,N_j\}$.}
\mychange{The constraints \eqnref{eq:milp_shc_big_M_1} and \eqnref{eq:milp_shc_big_M_2} 
  encode the switching decision for each of the switchable dependencies pairs
  using the big-M binary decision formulation \cite{Hult2015}.
Consider a $b \in \bvec$, if $b = 1$, \eqnref{eq:milp_shc_big_M_1} is
  relaxed because of the $-bM$ factor. 
Conversely, if $b = 0$, \eqnref{eq:milp_shc_big_M_2} is relaxed,
  because of the $-(1-b)M$ factor.
}

\subsection{Shrinking Horizon Feedback Control Scheme}

Having defined the \acs{MILP} in \eqnref{eq:milp_shc},
  we present an optimization-based 
  shrinking horizon feedback control scheme 
  to minimize the cumulative route completion times 
  of the \acs{AGVs} based on delays as they occur.
The scheme consists of an initial planning phase
  followed by an online phase.
During the planning phase, 
  the roadmap and \acs{AGV} start- and goal-positions are
  used to define a \acs{MAPF} problem.
The \acs{MAPF} problem is solved using an algorithm such as \acs{CBS}, \acs{ECBS},
  and the solution $\mathcal{P}$ is used to 
  construct an \acs{SADG} using \algref{alg:sadg_algorithm}.
Once constructed,
  the execution policy in \defref{def:seadg_execution} is used to navigate the
  the \acs{SADG}'s trivial solution, 
  $\SEADGgraph = \SADGgraph{\zerovec}$.
As \acs{AGVs} traverse the roadmap, potentially incurring delays, 
  the \acs{MILP} formulation in \eqnref{eq:milp_shc} is parameterized based 
  on the current \acs{AGV} route progress, and solved.
This solution is then used to update the \acs{SE-ADG} used by 
  the execution policy, until the next optimization iteration,
  where the \acs{MILP} is re-parameterized, and the \acs{SE-ADG} is updated.
This iterative loop repeats until all \acs{AGVs} reach their respective goals.
This scheme is illustrated in \figref{fig:shc_strategy_diagram}.

Having defined the feedback control scheme,
  we now prove that the feedback scheme is recursively feasible.
This implies that if the initial planning phase is completed,
  the \acs{MILP} will remain feasible until all \acs{AGVs} have reached their 
  respective goal positions. 
We use the notation $\bvec_{T_m}^*$ to refer to the minimizer 
  of \eqnref{eq:milp_shc}, parameterized by the \acs{AGV} positions 
   and solved at some time $t = T_m$, $m \in \mathbb{N}$.
\begin{myproposition}[Recursive Feasibility of \acs{SHC} scheme] 	
	If the initial \acs{SE-ADG}, $\SEADGgraph = \SADGgraph{\zerovec}$, 
	  obtained from the planning phase, is acyclic,
	  the shrinking horizon feedback control scheme
	  will guarantee that $\SADGgraph{\bvec^*}$ is acyclic
	  at each subsequent optimization step.
	\label{res:recursive_feasibility_shc}
\end{myproposition}
\begin{myproof}
	Proof by induction.
	Initially the \acs{SE-ADG}, $\SEADGgraph = \SADGgraph{\bvec^*_{T_0}}$ is acyclic.
	If the \acs{MILP} in \eqnref{eq:milp_shc} 
	  is solved at $t = T_1 > T_0$, 
	  \corref{res:acyclic_seadg_opt_has_solution}
	  guarantees that the \acs{SE-ADG} at $T_1$, $\SEADGgraph{}^{T_1}$, 
	  obtained from $\SADGgraph{\bvec^*_{T_1}}$, is acyclic.
	Subsequent optimization steps 
	  will always result in acyclic \acs{SE-ADG}s,
	  proving recursive feasibility of the feedback control scheme.
\end{myproof}
\begin{figure}[]
	\centering
	\begin{subfigure}{0.95\linewidth}
		\centering
		\includegraphics[width=\linewidth]{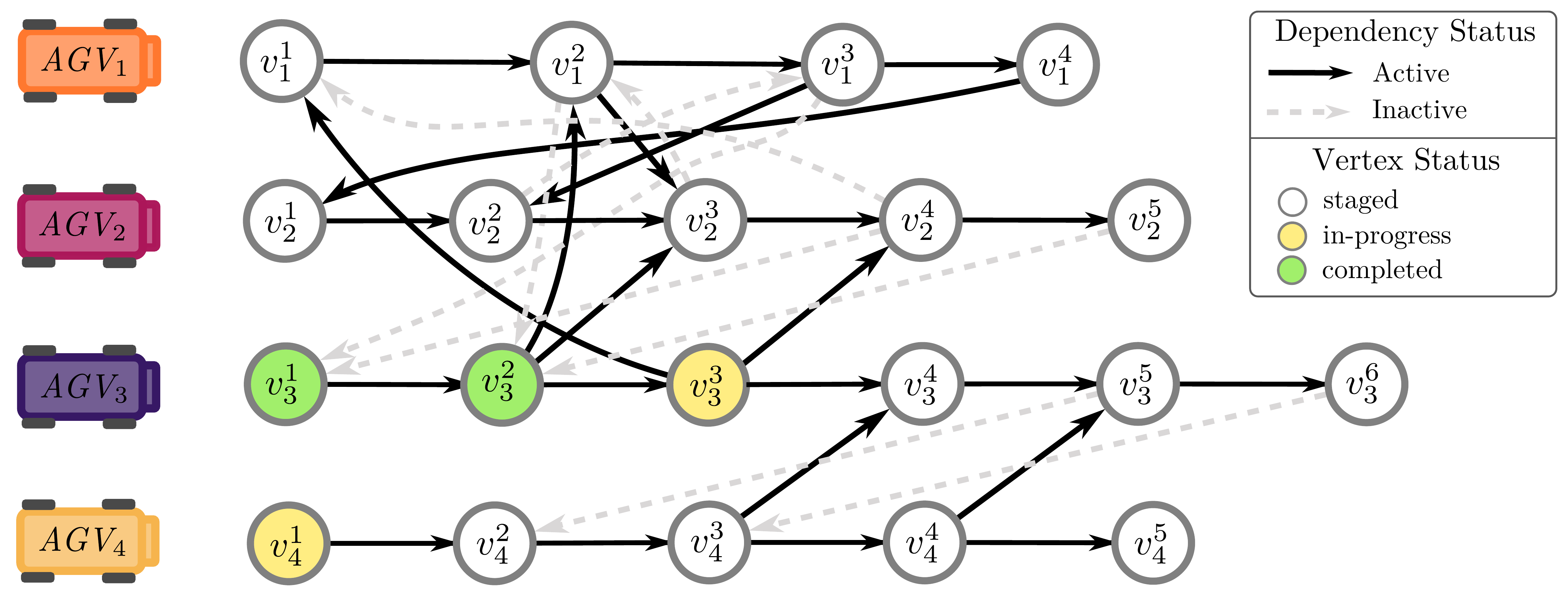}
		\caption{\acs{AGV} progress illustrated by vertex statuses. $\acs{AGV}_4$ is delayed.}
		\vspace{4mm}
		\label{fig:example_optimization_1}
	\end{subfigure}
	\hspace{4mm}
	\begin{subfigure}{0.95\linewidth}
		\centering 
		\includegraphics[width=\linewidth]{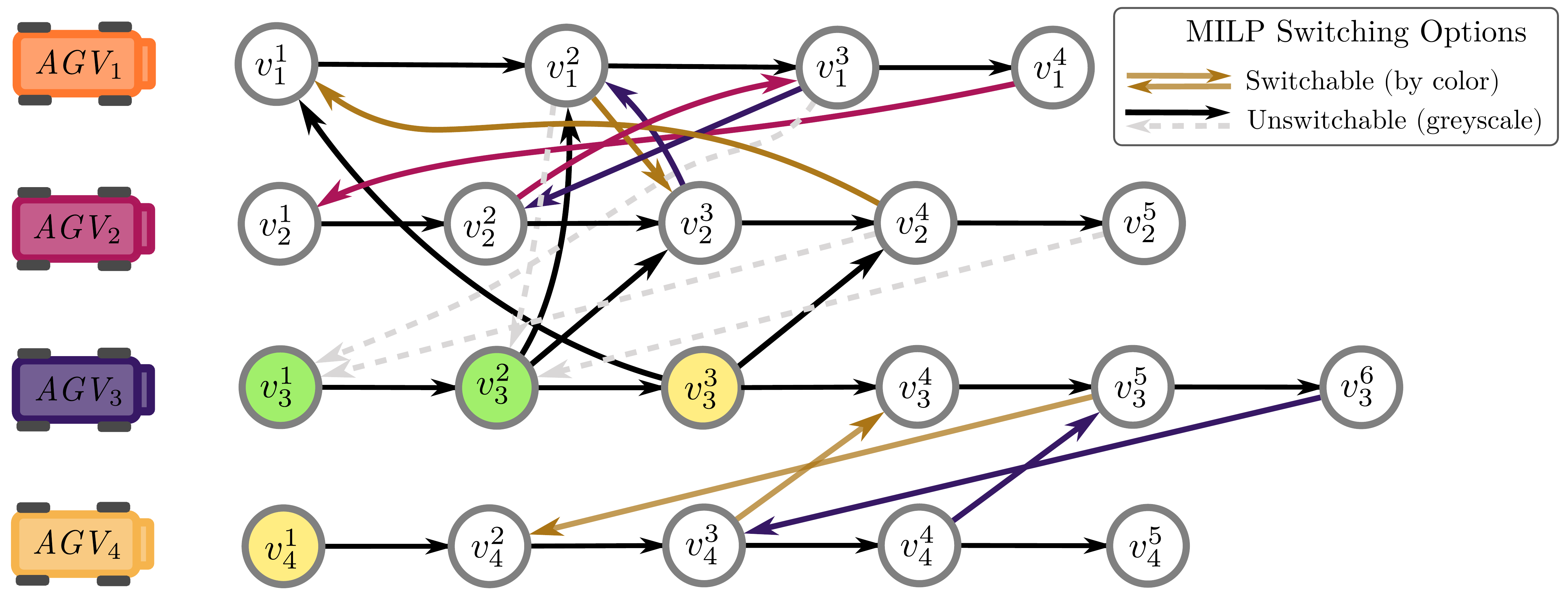}
		\caption{Based on \acs{AGV} progress, ordering can be adjusted by considering 
		the valid switching pairs within the \acs{MILP} formulation.}
		\vspace{4mm}
		\label{fig:example_optimization_2}
	\end{subfigure}
	\hspace{4mm}
	\begin{subfigure}{0.95\linewidth}
		\centering 
		\includegraphics[width=\linewidth]{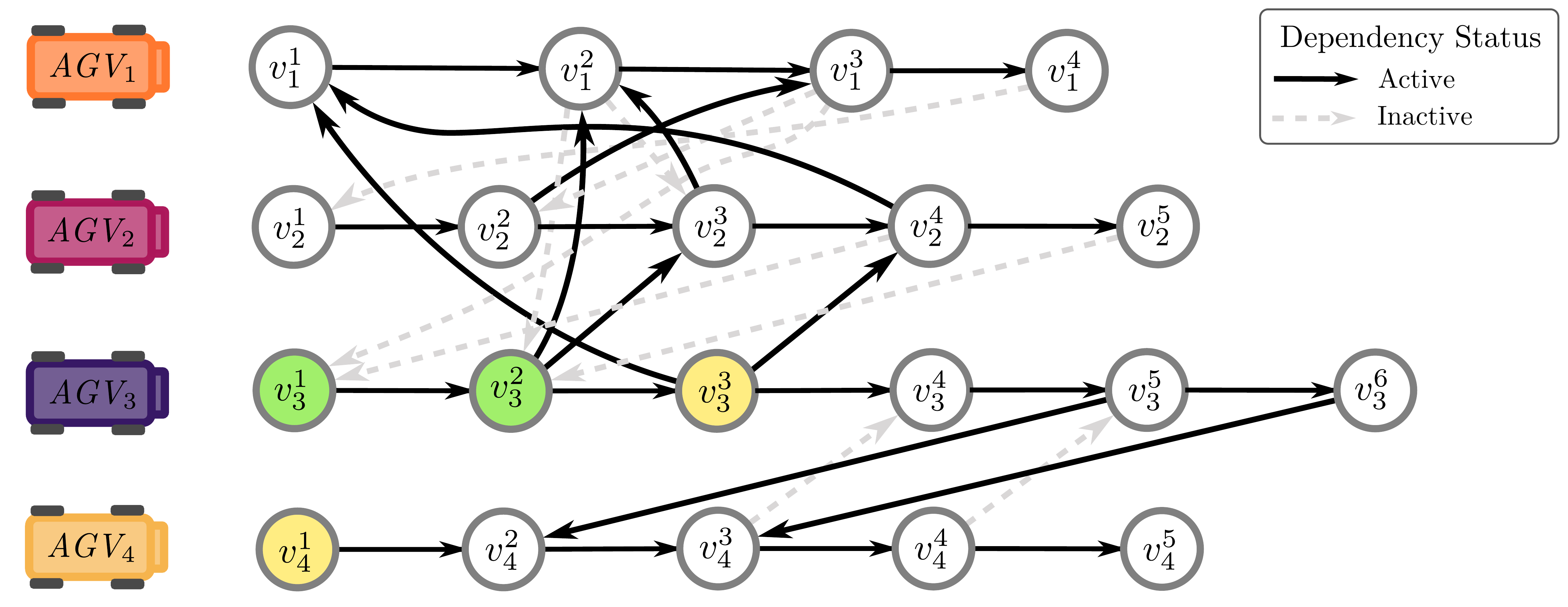}
		\caption{The switching enables $\acs{AGV}_3$ to continue without waiting for $\acs{AGV}_4$,
		and $\acs{AGV}_2$ without waiting for $\acs{AGV}_1$.}
		\label{fig:example_optimization_3}
	\end{subfigure}
	\caption{Illustrative example of switching performed by the shrinking horizon feedback control scheme.}
	\label{fig:example_optimization}
\end{figure}
To illustrate this feedback control scheme,
  an example of the online optimization 
  step is shown in \figref{fig:example_optimization}.
The example starts at time $t$, where $T_1 \leq t \leq T_2$,
  and $\acs{AGV}_4$ has been delayed.
However, because of the dependencies in 
  $\SADGgraph{\bvec^*_{T_1}}$,
  $\acs{AGV}_3$ must wait for $\acs{AGV}_4$ before it can proceed,
  as illustrated in \figref{fig:example_optimization_1}.
At $t = T_2$, the optimization step is performed
  and a new \acs{SE-ADG} is obtained, $\SADGgraph{\bvec^*_{T_2}}$,
  as illustrated in \figref{fig:example_optimization_2}. 
Note that $\SADGgraph{\bvec^*_{T_2}}$ has switched the dependencies 
  between $\acs{AGV}_3$ and $\acs{AGV}_4$, and remains acyclic.
For $t > T_2$, the \acs{AGVs} continue executing their plans,
  but using the newly obtained \acs{SE-ADG}, $\SADGgraph{\bvec^*_{T_2}}$.

\subsection{Grouping Switchable Dependency Pairs}
\label{sec:grouping}

\begin{figure}[h!]
	\centering
	\begin{subfigure}{1.0\linewidth}
		\centering
		\includegraphics[width=1.0\linewidth]{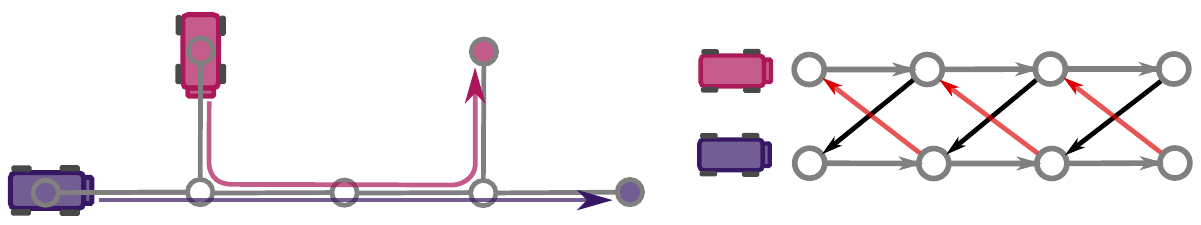}
		\caption{\acs{AGVs} travelling in the same direction and the corresponding \acs{SE-ADG}.}
		\vspace{4mm}
		\label{fig:dep_grouping_1}
	\end{subfigure}
	\begin{subfigure}{1.0\linewidth}
		\centering
		\includegraphics[width=1.0\linewidth]{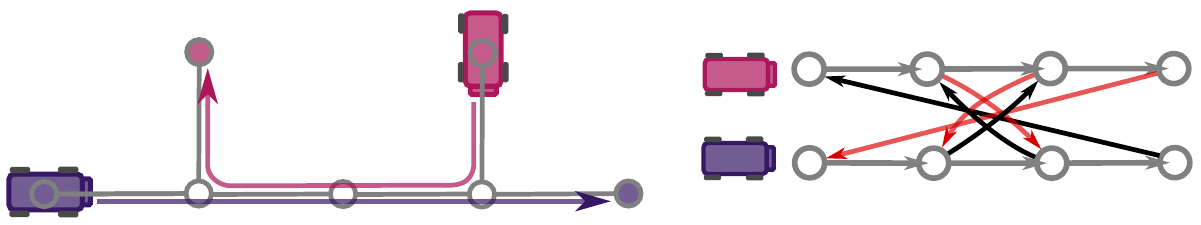}
		\caption{\acs{AGVs} travelling in the opposite direction and the corresponding \acs{SE-ADG}.}
		\label{fig:dep_grouping_2}
	\end{subfigure}
	\caption{Dependency grouping patterns for \acs{AGVs}
	planned to cross roadmap vertices in the (a) same and (b) opposite directions.}
	\label{fig:dep_grouping}
\end{figure}

Two patterns of switchable dependencies often occur with multi-agent 
  planning problems.
The first pattern is shown in \figref{fig:dep_grouping_1}
  and occurs when a \acs{MAPF} plan requires two \acs{AGVs} 
  to travel across multiple roadmap vertices in the same direction.
\mychange{
  This pattern can be found by sequentially visiting each inter-\acs{AGV}
  dependency $(v_i^k, v_j^l)$ 
  in the \acs{SADG} and searching for any sequence of 
    dependencies of the pattern 
  \begin{align*}
	(v_i^{k+n}, v_j^{l+n}), \; n \in {1,2,\dots}.
  \end{align*}
  A dependency group $DG$ then consists of the sequence 
  \begin{align*}
	DG = ((v_i^{k}, v_j^{l}), (v_i^{k+1}, v_j^{l+1}), (v_i^{k+2}, v_j^{l+2}), \dots ).
  \end{align*}
}

  Similarly, the second pattern is shown in \figref{fig:dep_grouping_2},
  corresponding to two \acs{AGVs} travelling in the opposite direction.
\mychange{
	This pattern can be found by sequentially visiting each inter-\acs{AGV}
	dependency $(v_i^k, v_j^l)$ 
	in the \acs{SADG} and searching for any sequence of 
	  dependencies of the pattern 
	\begin{align*}
	  (v_i^{k+n}, v_j^{l-n}), \; n \in {1,2,\dots} .
	\end{align*}
	A dependency group $DG$ then consists of the sequence 
	\begin{align*}
	  DG = ((v_i^{k}, v_j^{l}), (v_i^{k+1}, v_j^{l-1}), (v_i^{k+2}, v_j^{l-2}), \dots ).
	\end{align*}
}

In both cases,
  a single binary variable is sufficient to express 
  the switching for all the dependency pairs in the group,
  since the only way for the switching to yield an acyclic 
  \acs{SE-ADG} is if either all the forward dependencies
  and none of the reverse dependencies are active (or vice-versa).
Depending on the roadmap topography, 
  the size of $\bvec$ can be significantly reduced, 
  greatly reducing the complexity of the \acs{MILP} problem 
  at each iteration.
\mychange{
  Identifying dependency groups is an $\mathcal{O}(n)$ operation, 
  and can be done during \acs{SADG} construction, i.e., before plan 
  execution.
}

\subsection{Summary}

The result of this section is that
  we have an \acs{SADG} from which an \acs{OCP} is derived which 
  can be solved at any time-step to re-order the \acs{AGVs} 
  based on their progress of the plans.
A feasible solution is guaranteed to yield a deadlock-free, collision 
  free plan for the AGVs.

%% file: sec/6_rhc.tex

\section{Receding Horizon Control Scheme}
\label{sec:rhc}

We now address the fact that the 
  OCP in \eqnref{eq:ocp_shc} 
  could yield an optimization problem too large for feasible 
  real-time implementation as it considers the entire,
  finite-length plans.
Specifically, 
  we show how the \acs{OCP} in \eqnref{eq:ocp_shc}
  can be approximated by a receding horizon variant 
  of the \acs{MILP} in \eqnref{eq:milp_shc}.
The motivation for this is predicated on the fact that 
  \acs{MILP} problems are exponentially complex 
  in the number of discrete 
  variables.
Furthermore, a receding horizon formulation allows
  for ad-hoc re-planning of \acs{AGVs} without needing to 
  wait for all \acs{AGVs} to reach their goals.
To this end, 
  we first introduce a method to split an \acs{SADG}
  into a smaller \acs{SADG} and an acyclic \acs{SE-ADG}.
We then show how re-formulating the \acs{MILP}
  to consider the smaller \acs{SADG} approximates
  a solution to the original \acs{OCP}
  and maintains recursive feasibility guarantees.

Note that we extend the 
  approach originally presented in \cite{berndtFeedbackSchemeSADGICAPS2020}
  by introducing a method to split the \acs{SADG}.
This enables the persistent re-solving of the \acs{MAPF}
  mid-route as shown for the original \acs{ADG} 
  method in \cite{hoenigPersistentRobustExecution2018}.

\subsection{Introducing The Finite Horizon \acs{SADG} Subset}

\acf{RHC} approaches are predicated on the fact that 
  a sufficiently accurate solution to an \acs{OCP} can be 
  obtained by only considering the system trajectories 
  within a finite horizon.
For discrete systems, such as the one described by a \acs{SADG},
  defining the finite horizon which guarantees the \acs{RHC} 
  presents a unique challenge: 
  re-ordering \acs{AGVs} within a finite horizon
  could still yield a cyclic \acs{SE-ADG}, despite the portion of the 
  \acs{SE-ADG} within the horizon being acyclic.

To this end, 
  we present \algref{alg:fh_sadg} to split an \acs{SADG}
  into 1.) a \textit{finite horizon \acs{SADG}} 
  and 2.) an \textit{acyclic \acs{SE-ADG}}
  such that if the \textit{finite horizon \acs{SADG}} (a subset of the
  original \acs{SADG}) yields an acyclic, finite horizon \acs{SE-ADG}
  for a particular $\bvecFH \subset \bvec$,
  the entire \acs{SE-ADG} will be acyclic.
This \textit{finite horizon \acs{SADG}} can then be used, instead of the 
  original \acs{SADG}, to parameterize \acs{MILP} formulation,
  greatly reducing the computational load and maintaining 
  collision- and deadlock-free plan execution guarantees.

\begin{figure}[]
	\centering
	\includegraphics[width=0.95\linewidth]{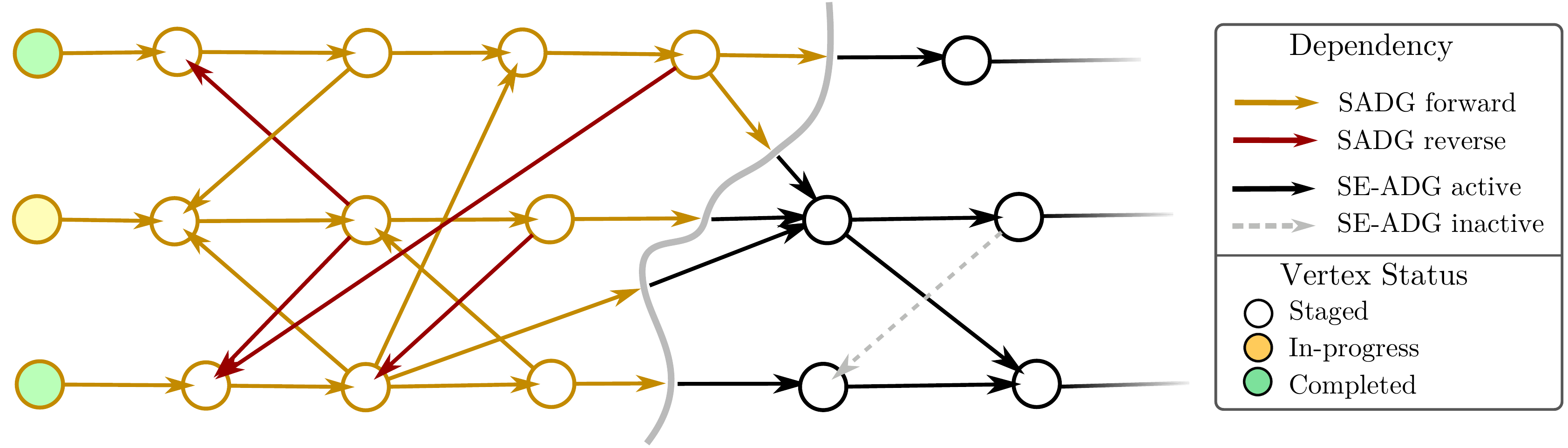}
	\caption{Graphical illustration of how the original \acs{SADG} can
	be split into a finite horizon \acs{SADG} subset and an \acs{SE-ADG}, while ensuring 
	if the finite horizon \acs{SADG} subset yields an acyclic \acs{SE-ADG} subset,
	the resultant \acs{SE-ADG} will be acyclic.}
	\label{fig:finite_horizon_sadg_subset}
\end{figure}

To this end, 
  we introduce \lemref{res:sadg_split},
  a result which enables the splitting 
  of an \acs{SADG} into 
  a finite horizon \acs{SADG} subset and an \acs{SE-ADG} subset,
  the resulting \acs{SE-ADG} will be acyclic 
  if the switching within the finite horizon \acs{SADG} subset yields an 
  acyclic \acs{SE-ADG}.
\lemref{res:sadg_split} builds on the more general fact that 
  a graph $\mathcal{G}$ is acyclic if 1) it is constructed from 
  two acyclic graphs, $\mathcal{G}_1 = (\mathcal{V}_1, \mathcal{E}_1)$ 
  and $\mathcal{G}_2 = (\mathcal{V}_2, \mathcal{E}_2)$,
  and 2) all edges connecting $\mathcal{G}_1$ and $\mathcal{G}_2$ 
  point from $\mathcal{V}_1$ to $\mathcal{V}_2$.
This result is proven in Appendix A, \lemref{res:two_acyclic_graphs}.
The result in \lemref{res:sadg_split} is 
  illustrated by the example in \figref{fig:finite_horizon_sadg_subset}.
The grey partition line splits the \acs{SADG}
  into a \acs{SADG} subset and an \acs{SE-ADG}.
Note how all the dependencies connecting the \acs{SADG} subset
  and the \acs{SE-ADG} go from the \acs{SADG} subset to the \acs{SE-ADG}.

\begin{figure}[]
	\centering
	\includegraphics[width=0.9\linewidth]{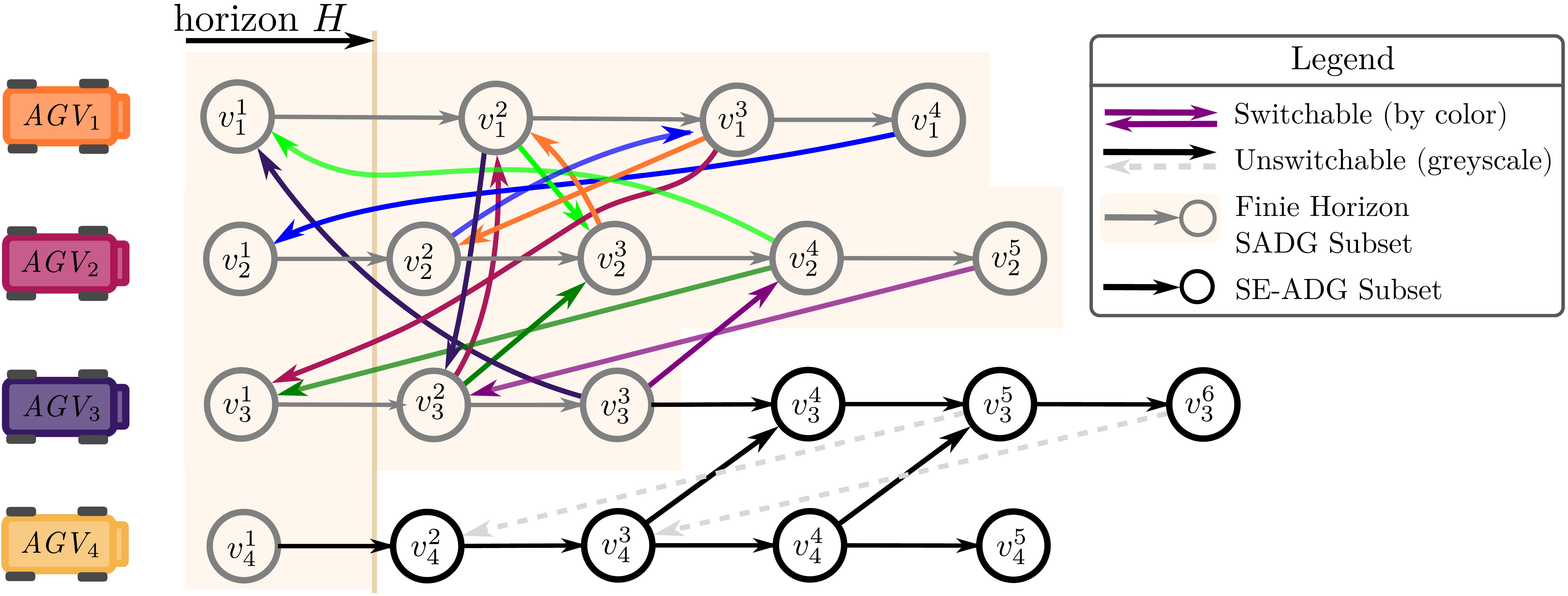}
	\caption{Highlighted finite horizon \acs{SADG} subset connected to the acyclic
	\acs{SE-ADG} subset with unidirectional dependencies. 
	The horizon $H$ is indicated by the horizontal black arrow.}
	\label{fig:finite_horizon_sadg_subset_split}
\end{figure}

The application of \algref{alg:fh_sadg} 
  to our running example is illustrated in \figref{fig:finite_horizon_sadg_subset_split}.
Here, the finite horizon \acs{SADG} subset is highlighted in orange.

\begin{algorithm}[]
\caption{Extracting a Finite Horizon \acs{SADG} Subset}
\small
\begin{algorithmic}[1]
	\renewcommand{\algorithmicrequire}{\textbf{Input:}}
	\renewcommand{\algorithmicensure}{\textbf{Result:}}
	\REQUIRE $\SADGgraph{\bvec}$, $\horizon$
	\ENSURE $\SADGgraphFH{\bvecFH}$ \\
	\vspace{2mm}
	// Add switching dependency pairs within horizon $\horizon$
	\STATE Add vertices to $\SADGverticesFH$ within $\horizon$ \label{alg:fh_sadg_start}
	\STATE Add all edges to/from $\SADGverticesFH$ to $\SADGedgesFH$ 
	\STATE Add $b \in \bvec$ to $\bvecFH$ if $b$ is related to edges in $\SADGedgesFH$
	\STATE Add all edges pointing to $v \in \SADGverticesFH$ to $\mathcal{E}_\text{inwards}$ \label{alg:fh_sadg_end} \\
	\vspace{2mm}
	// Expand $\SADGverticesFH$ until no dependencies point into $\SADGverticesFH$
	\WHILE {$\mathcal{E}_\text{inwards} \neq \emptyset$} \label{alg:fh_sadg_start_while}
		\STATE $e \leftarrow \text{pop}(\mathcal{E}_\text{inwards})$
		\STATE Add all $v$'s pointing to $e$
		\STATE Add all edges pointing from $v$ out of $\SADGverticesFH$ to $\mathcal{E}_\text{inwards}$
	\ENDWHILE \label{alg:fh_sadg_end_while}
	\vspace{1mm}
	\RETURN $\SADGgraphFH{\bvecFH} = \big( \SADGverticesFH, \SADGedgesFH(\bvecFH) \big)$
\end{algorithmic} 
\label{alg:fh_sadg}
\end{algorithm}

\begin{figure}[]
	\centering
	\begin{subfigure}{0.48\textwidth}
		\begin{minipage}{0.08\linewidth}
			\caption{ }
			\label{fig:select_sadg_subset_1}
		\end{minipage}
		\hfill 
		\begin{minipage}{0.9\linewidth}
			\includegraphics[width=\linewidth]{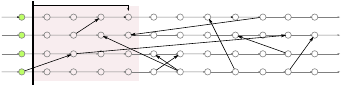}
		\end{minipage}
	\end{subfigure}
	\begin{subfigure}{0.48\textwidth} 
		\begin{minipage}{0.08\linewidth}
			\caption{ }
			\label{fig:select_sadg_subset_2}
		\end{minipage}
		\hfill 
		\begin{minipage}{0.9\linewidth}
			\includegraphics[width=\linewidth]{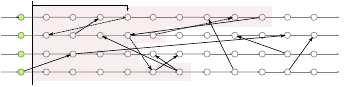}
		\end{minipage}
	\end{subfigure}
	\begin{subfigure}{0.48\textwidth}
		\begin{minipage}{0.08\linewidth}
			\caption{ }
			\label{fig:select_sadg_subset_3}
		\end{minipage}
		\hfill 
		\begin{minipage}{0.9\linewidth}
			\includegraphics[width=\linewidth]{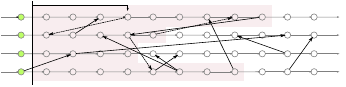}
		\end{minipage}
	\end{subfigure}
	\begin{subfigure}{0.48\textwidth}
		\begin{minipage}{0.08\linewidth}
			\caption{ }
			\label{fig:select_sadg_subset_4}
		\end{minipage}
		\hfill 
		\begin{minipage}{0.9\linewidth}
			\includegraphics[width=\linewidth]{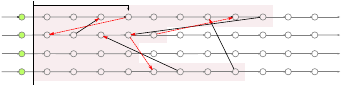}
		\end{minipage}
	\end{subfigure}
	\caption{Illustrative example of how \algref{alg:fh_sadg} 
	  extracts the finite horizon \acs{SADG} subset 
	  from the original \acs{SADG} 
	  for a user-specified horizon $H$.
	  \mychange{
		The magenta region refers to the stack $\SADGverticesFH$ in \algref{alg:fh_sadg}.
		(a) refers to line 1, with the vertices within $H$ added to $\SADGverticesFH$.
		(b) refers to line 2, where all \textbf{forward or reverse} dependencies 
		pointing to within $H$ are added 
		to $\SADGverticesFH$,
		 and (c) to line 3-4, where any \textbf{forward} dependencies 
		 pointing to $\SADGverticesFH$ in (b) are
		 included.
		(d) shows the switchable dependencies, in red, 
		as identified in lines 5-8.
	  }
	 }
	\label{fig:select_sadg_subset}
\end{figure}

Extracting a finite horizon \acs{SADG} subset 
  from the \acs{SADG} can be done 
  with \algref{alg:fh_sadg}.
\mychange{
To illustrate the intuition behind \algref{alg:fh_sadg},
  consider \figref{fig:select_sadg_subset}. 
In line 1, \algref{alg:fh_sadg} navigates the graph $\SADGgraph{\bvec}$
  by first pushing all the vertices estimated to be completed 
  within a user-specified time horizon $H$ 
  to a stack $\SADGverticesFH$, shown in \figref{fig:select_sadg_subset_1}.
Next, each forward-reverse dependency pair pointing to a vertex 
  in $\SADGverticesFH$ is identified,
  and its associated binary variable is appended to $\bvecFH$.
This is illustrated by the magenta region in \figref{fig:select_sadg_subset_2}.
The remaining forward dependencies pointing to vertices 
  in $\SADGverticesFH$ are appended to $\SADGedgesFH$,
  until no dependencies point from a vertex in $\SADGverticesFH$
  to a vertex outside $\SADGverticesFH$.
  This is shown in \figref{fig:select_sadg_subset_3}
The resulting switchable dependencies 
  which are considered within $\bvecFH$ are shown in \figref{fig:select_sadg_subset_4}.
}


\begin{mylemma}[Finite Horizon \acs{SADG} Subset solution guarantees 
	acyclicity of resultant \acs{SE-ADG}]
	Consider a \acs{SADG}, $\SADGgraph{\bvec}$, split into 
	a finite horizon \acs{SADG} subset and an \acs{SE-ADG} subset
	using \algref{alg:fh_sadg}.
	If $\bvecFH^*$ is such that $\SADGgraphFH{\bvecFH^*}$ is acyclic,
	the resultant \acs{SE-ADG} is acyclic.
  \label{res:sadg_split}
\end{mylemma}
\begin{myproof}
	By lines \ref{alg:fh_sadg_start_while}-\ref{alg:fh_sadg_end_while}, 
	\algref{alg:fh_sadg} ensures all dependencies 
	connecting the finite horizon \acs{SADG} subset, $\SADGgraphFH{}$,
	and the \acs{SE-ADG} subset, $\SEADGgraphSubset$, 
	are directed from $\SADGgraphFH{}$ to $\SEADGgraphSubset$ only.
	Using the result in \lemref{res:two_acyclic_graphs},
	  if a particular $\bvecFH$ is chosen such that 
	  $\SADGgraphFH{\bvecFH}$ is acyclic, 
	  the resultant \acs{SE-ADG} will be acyclic,
	  since $\SEADGgraphSubset$ is acyclic as well.
\end{myproof}

\subsection{Re-formulation of \acs{MILP}}

We now re-formulate the \acs{MILP} to consider the finite horizon \acs{SADG} subset 
  instead of the entire \acs{SADG} as was done in \eqnref{eq:milp_shc}.
\begin{subequations}
	\begin{align}
		& \min_{\bvec, \tgvec, \tsvec} \sum_{i=1}^{N} t_g(v_i^{n_i}) \label{eq:milp_rhc_cost_fun} \\
		& \hspace{5mm} \text{s.t.} \nonumber \\ 
		t_g(v_i^k) &> t_s(v_i^k) + \Delta T(v_i^k) \; \forall \; v_i^k \in \Pi \big( \SADGverticesFH ({\bvec}) \big), \label{eq:milp_rhc_1_1} \\ 
		t_s(v_i^{k+1}) &> t_g(v_i^k)   \; \forall \; v_i^{k+1} \in \Gamma \big( \SADGverticesFH ({\bvec}) \big), \label{eq:milp_rhc_1_2} \\
		\status{v'} &= \staged \;\; \forall \; (v,v') \; \in \Psi(\bvec, \bvecstart, \SADGverticesFH ({\bvec}) ), \label{eq:milp_rhc_2_1} \\
		t_g(v_i^k) &< t_s(v_j^l) \;\; \forall \; (v_i^k, v_j^k) \in \SADGedgesFH ({\bvec}) \; \text{if} \;\; i \neq j,  \label{eq:milp_rhc_2_2} \\
		t_s(v_j^l) &> t_g(v_i^k) - b M, \; \forall \; \; b \in \bvec \\
		t_s(v_i^{k-1}) &> t_g(v_{j}^{l+1}) - (1 - b) M, \; \forall \; \; b \in \bvec 
	\end{align}
\label{eq:milp_rhc}
\end{subequations}
where $M \gg 0$ is a large constant 
  \mychange{
  and $n_i$ is the horizon length as determined for \acs{AGV} $i$ 
  as determined by \algref{alg:fh_sadg}.
} 

\begin{figure*}[]
	\centering
	\includegraphics[width=0.95\textwidth]{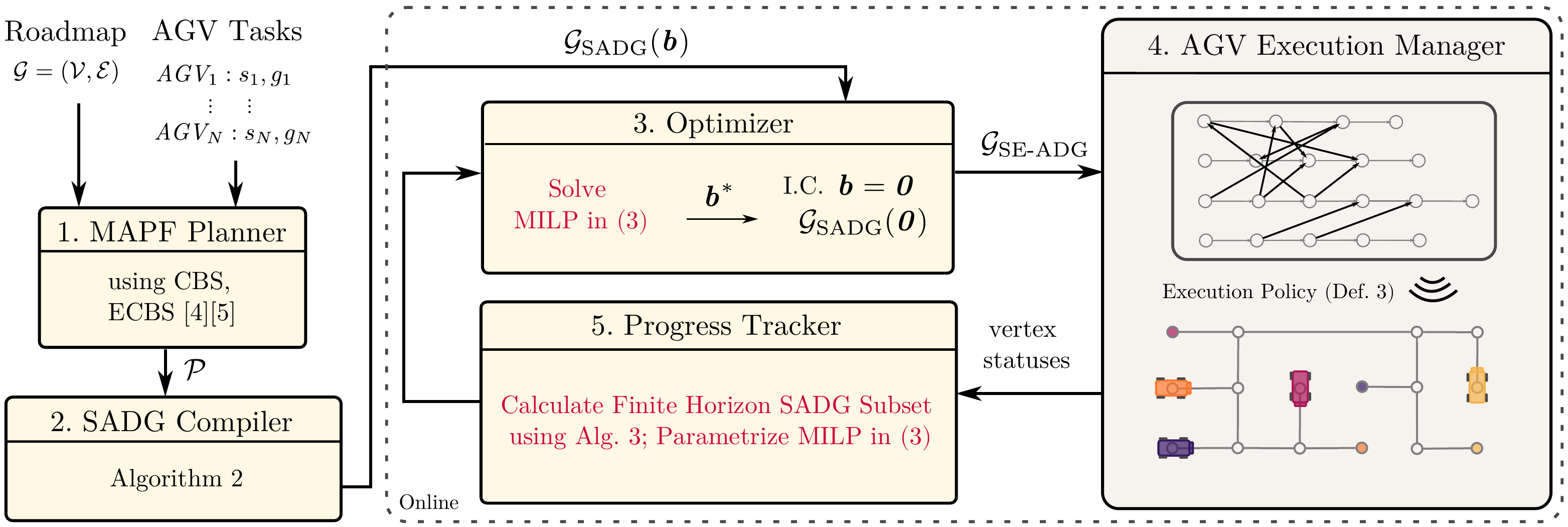}
	\vspace{2mm}
	\caption{\textbf{Receding horizon feedback control diagram:} 
	\textit{Differences from \figref{fig:shc_strategy_diagram} 
	are highlighted in red.}
	As \acs{AGVs} progress,
	    and are inevitably delayed, their status is tracked, 
	  and an optimization 
		iteration is performed to re-order the \acs{AGVs} 
		based on a Finite Horizon \acs{SADG} Subset 
		calculated by \algref{alg:fh_sadg}.
	  The solution of the \acs{MILP} is used to update 
	    the \acs{SE-ADG}
		used for plan execution in a feedback loop.
	}
	\label{fig:rhc_strategy_diagram}
\end{figure*}

\subsection{Receding Horizon Feedback Control Scheme}

Having re-formulated the \acs{MILP} in \eqnref{eq:milp_rhc}
  such that a subset of the \acs{SADG} 
  is considered at each optimization step,
  we present the
  receding horizon feedback control scheme as 
  shown in \figref{fig:rhc_strategy_diagram}.
The main difference 
  between the \acs{RHC} scheme 
  and the \acs{SHC} scheme presented in \secref{sec:shc}
  is the inclusion of \algref{alg:fh_sadg} in the step \textit{5. Track \acs{AGV} Progress}.
Finally,
  we prove recursive feasibility of the \acs{RHC} scheme.

\begin{myproposition}[Recursive Feasibility of \acs{RHC} scheme]
	If the initial \acs{SE-ADG}, $\SEADGgraph = \SADGgraph{\zerovec}$, 
		obtained from the planning phase, is acyclic,
		the \acl{RHC} scheme
		will guarantee that $\SADGgraph{\bvec^*}$ is acyclic
		at each subsequent optimization step.
	\label{res:rhc_recursive_feasibility}
\end{myproposition}
\begin{myproof}
	Proof by induction.
	Initially the \acs{SE-ADG}, $\SEADGgraph = \SADGgraph{\bvec^*_{T_0}}$ is acyclic.
	\algref{alg:fh_sadg} is used to extract the 
	  finite horizon \acs{SADG} subset at $t = T_1 > T_0$.
	If the \acs{MILP} in \eqnref{eq:milp_rhc} 
	  is solved at $t = T_1 > T_0$, 
	  \corref{res:acyclic_seadg_opt_has_solution}
	  guarantees that the \acs{SE-ADG} at $T_1$, $\SEADGgraph{}^{T_1}$, 
	  obtained from $\SADGgraph{\bvec^*_{T_1}}$, is acyclic.
	Subsequent optimization steps 
	  will always result in acyclic \acs{SE-ADG}s,
	  proving recursive feasibility of the feedback control scheme.
\end{myproof}

%% file: sec/7_alternative_solvers.tex

%% file: sec/8_eval.tex
\section{Evaluation}
\label{sec:eval}

We perform extensive evaluations of our proposed
  feedback control scheme.
Our simulations consist of multiple \acs{AGVs}
  occupying various roadmaps with randomized 
  start and goal locations.
Each simulation starts with a planning 
  step where the \acs{MAPF} is solved given the 
  start and goal locations.

To gain insight into our proposed control schemes, 
  we perform evaluations in three different 
  settings.
In the first, we aim to gain statistical insight into 
  the performance gains of our approach compared to 
  the original \acs{ADG} baseline
  by considering various 
  roadmap topologies and \acs{AGV} fleet sizes.
In the second setting,
  we compare our method to the state-of-the-art 
  robust \acs{MAPF} planner \Kalg \cite{ChenAAAI21b}.
In the third setting, we showcase our method 
  in a high-fidelity \acs{ROS} and Gazebo simulation environment.
\mychange{In all three settings, the \acs{RHC} \acs{SADG}
  is used in combination with dependency grouping as described in 
  \secref{sec:grouping}.}
All simulations were conducted on a Lenovo Thinkstation 
  with an Intel\textregistered{} Xeon 
  E5-1620 3.5GHz processor and 64 GB RAM.

\subsection{Setting 1: Statistical Analysis}
\label{sec:eval-setting-1}

We compare our proposed 
  receding horizon feedback control scheme,
  referred to from here on as the \acs{RHC} \acs{SADG} approach, 
  to the original \acs{ADG} approach 
  in \cite{hoenigPersistentRobustExecution2018}.
We consider the roadmaps in \figref{fig:roadmaps},
  inspired by intralogistic warehouse 
  layouts \cite{wurmanCoordinatingHundredsCooperative2008, vda5050, rmf_framework}. 
For each simulation run, 
  \acs{AGVs} are given randomized start and goal locations
  on the map. 
The execution policy in \defref{def:seadg_execution}
  is used for both the \acs{ADG} and \acs{RHC} \acs{SADG} 
  approaches.
The \acs{AGVs} are simulated as simple differential-drive 
  robots with constant rotational and translational velocities
  of $3 \rads$ and $1 \ms$ respectively.
The coordination of \acs{AGVs} is performed using \acs{ROS},
  where a central coordination \textit{\acs{ROS} node} 
  solves the \acs{MAPF}, constructs the \acs{RHC} \acs{SADG} (or \acs{ADG}), 
  and communicates the vertex statuses to each 
  \acs{AGV}.
The feedback loop sampling time $h$ is set to $2 \textit{ s}$.

\acs{AGVs} are artificially delayed as follows:
  at the start of each interval of length $\delaytime$, 
  $20\%$ of the \acs{AGV} fleet is \textit{randomly selected} and  
  stopped for the next $\delaytime$ seconds.
For the next interval, a different subset of the \acs{AGVs}
  is randomly selected and delayed, and so on.
Improvement is quantified by comparing the 
  cumulative route completion time of all the \acs{AGVs}
  for the same \acs{MAPF} plan when using the 
  baseline \acs{ADG} approach to our receding horizon 
  feedback control scheme,
  defined as 
\begin{align}
  \textit{improvement} = \frac{\sum t_\textit{ADG} - \sum t_\textit{\acs{RHC} \acs{SADG}}}{\sum t_\textit{ADG}} \cdot 100 \%,
  \label{eq:improvement}
\end{align}
where $\sum t_\ast$ 
  refers to the cumulative plan completion 
  time for all \acs{AGVs} using approach $\ast$.

\subsubsection*{Improvement and Delays}
\begin{figure}[]
  \centering 
  \includegraphics[width=\linewidth]{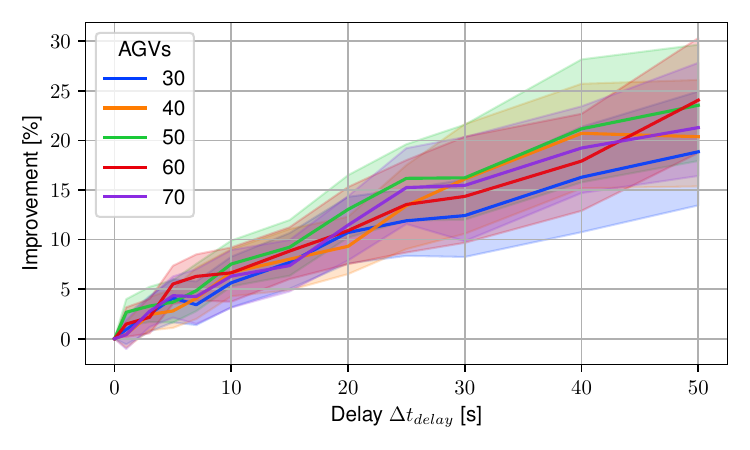}
  \caption{
    Improvement 
      for various simulated \acs{AGV} delays navigating  
      roadmaps in \figref{fig:roadmaps}.
    Each $\delaytime$ seconds, a randomly selected subset of $20\%$ 
      of the \acs{AGVs} are stopped for $\delaytime$ seconds.
    Horizon $\horizon = 5$.
    Bounds correspond to $\pm 1 \sigma$.
  }
  \vspace{-5mm}
  \label{fig:eval_improv_vs_delay}
\end{figure}
\begin{figure*}[]
  \centering
  \begin{subfigure}[h]{0.245\linewidth}
    \centering
    \includegraphics[width=\textwidth]{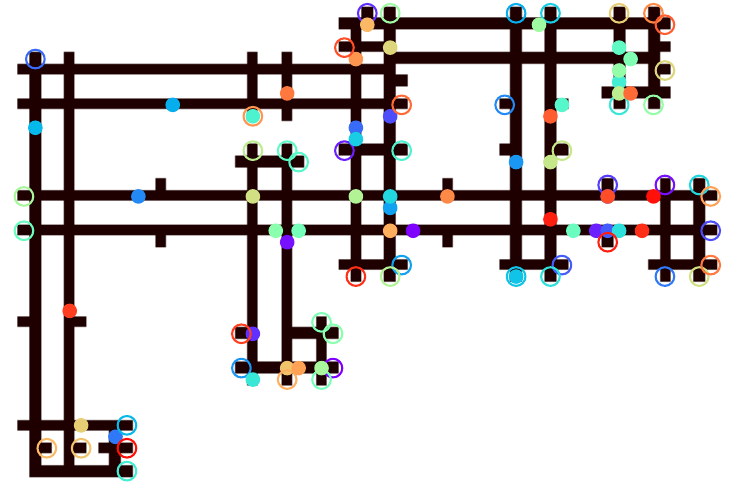}
    \vspace{0mm}
    \caption{Warehouse}
    \label{fig:roadmaps_0}
  \end{subfigure}
  \begin{subfigure}[h]{0.245\linewidth}
    \centering
    \includegraphics[width=\textwidth]{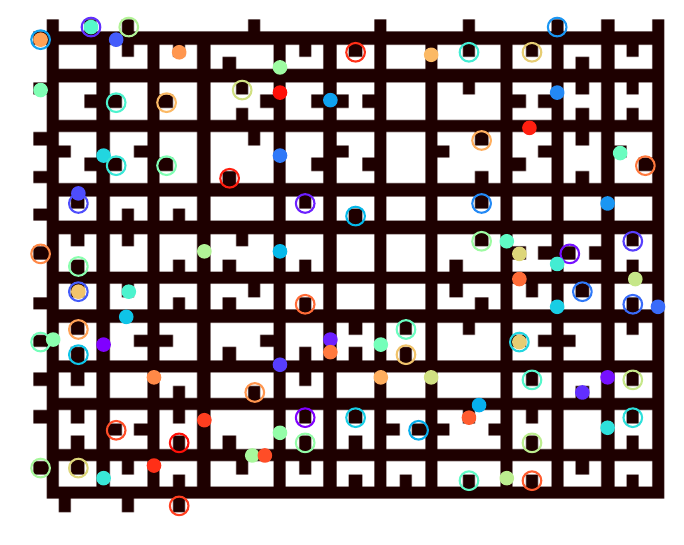}
    \caption{Full Maze}
    \label{fig:roadmaps_1}
  \end{subfigure}
  \begin{subfigure}[h]{0.245\linewidth}
    \centering
    \includegraphics[width=\textwidth]{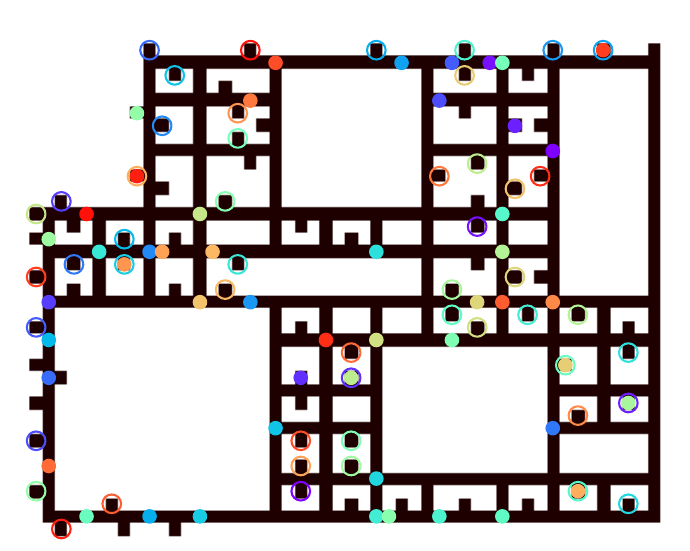}
    \caption{Half Maze}
    \label{fig:roadmaps_2}
  \end{subfigure}
  \begin{subfigure}[h]{0.245\linewidth}
    \centering
    \includegraphics[width=\textwidth]{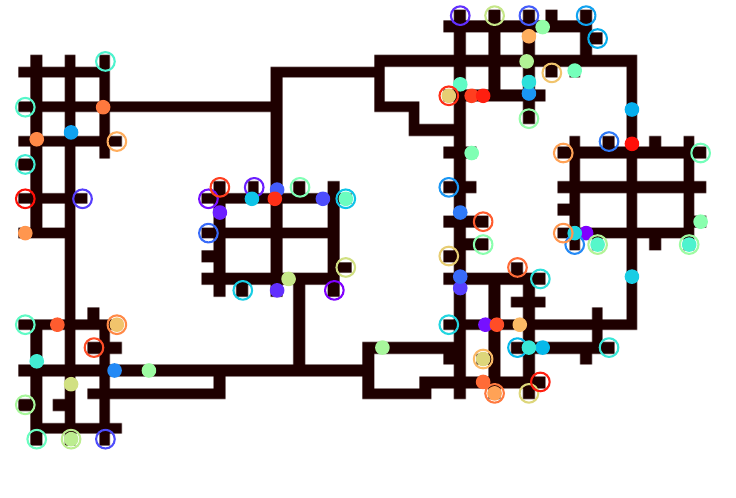}
    \vspace{0mm}
    \caption{Islands}
    \label{fig:roadmaps_3}
  \end{subfigure}
  \caption{Roadmaps used for the statistical analysis in \secref{sec:eval-setting-1}, 
    inspired by \cite{wurmanCoordinatingHundredsCooperative2008, vda5050, rmf_framework}.
    \acs{AGVs} are indicated by colored circles, and their corresponding goal location 
    by the same colored ring. Roadmap vertices are represented by black squares.
  }
  \label{fig:roadmaps}
\end{figure*}

We consider delays of $\delaytime \in \{ 1,3,5,10,15,20,25,30,40,50 \}$ seconds
  for \acs{AGV} fleet sizes of $\{30,40,50,60,70\}$
  navigating the four roadmaps in \figref{fig:roadmaps}.
We run $100$ simulations for each fleet size and delay time permutation.
For each simulation, the \acs{MAPF} is solved using \acs{ECBS} 
  with sub-optimality bound $w$ chosen such that the initial planning 
  time is below ten minutes.
\figref{fig:eval_improv_vs_delay} shows 
  the improvement for various \acs{AGV} fleet sizes 
  and delays.
The horizon $\horizon$ is set to $5$ seconds.
We observe that the improvement is almost linear with respect to the 
  delays of the \acs{AGVs}, for all the considered fleet sizes.
The improvement standard deviation, $\sigma$, indicated by the lightly 
  colored bands, is relatively small, indicating that improvement is 
  consistent for a given fleet size and delay configuration.
\subsubsection*{Improvement and Horizon}
\begin{figure}[]
  \centering
  \begin{subfigure}[h]{\linewidth}
    \centering
    \includegraphics[width=\textwidth]{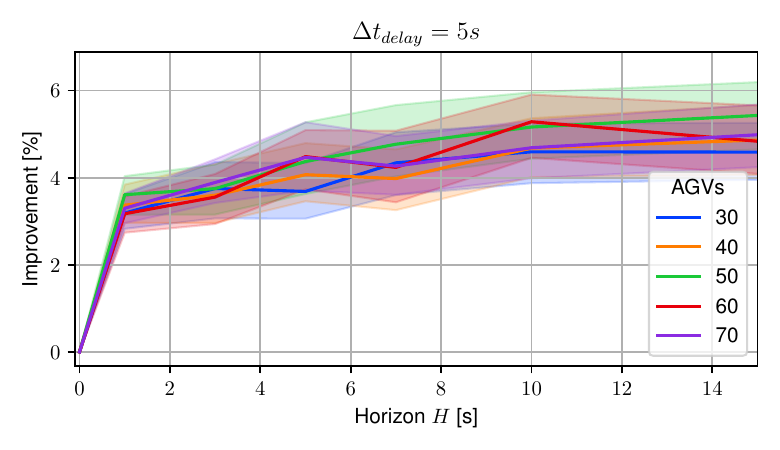}
  \end{subfigure}
  \begin{subfigure}[h]{\linewidth}
    \centering
    \includegraphics[width=\textwidth]{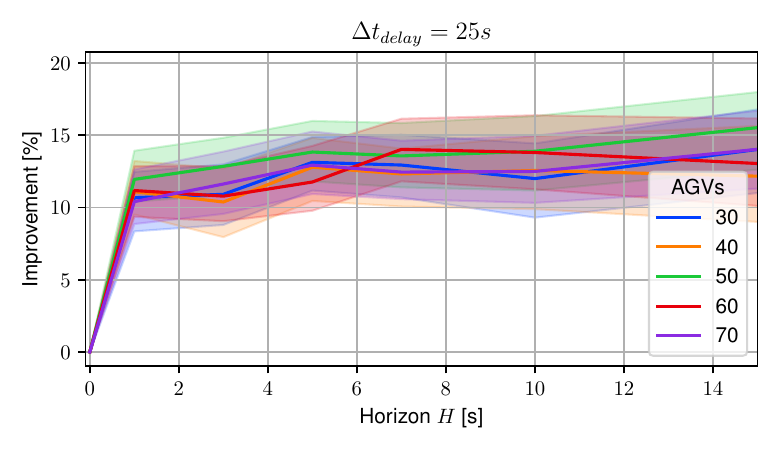}
  \end{subfigure}
  \caption{
    Improvement for delays $\delaytime$ of $5$ and $25$ 
      seconds given various horizons $\horizon{}$.
      Bounds correspond to $\pm 1 \sigma$.
  }
\label{fig:horizon_improv}
\end{figure}

For the \acs{RHC} \acs{SADG} approach, 
  we consider various horizons $\horizon \in \{ 1, \dots, 15 \}$ s. 
We consider delays $\delaytime = 5$s
  and $\delaytime = 25$s, with the improvement shown 
  in \figref{fig:horizon_improv}.
We run $100$ simulations for each fleet size, horizon and delay time permutation.
Notice how for both shorter and longer delays,
  the improvement already significantly increases
  with low horizons $\horizon$,
  indicating that the \acs{RHC} \acs{MILP} in \eqnref{eq:milp_rhc}
  performs a consistently good approximation of the 
  \acs{OCP} in \eqnref{eq:ocp_shc} for small $\horizon$.

\begin{figure}[]
  \centering 
  \includegraphics[width=\linewidth]{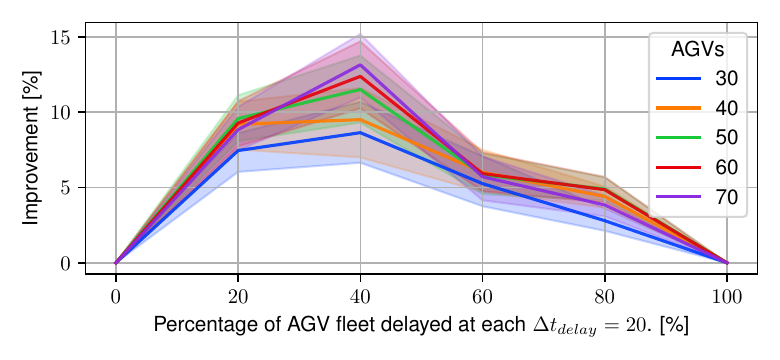}
  \caption{
    Improvement for various percentages of \acs{AGVs} delayed at each 
    interval of $\delaytime{}$. 
    $\delaytime{} = 20$s and horizon $\horizon = 5$s.
    Bounds correspond to $\pm 1 \sigma$.
  }
  \label{fig:eval_improv_vs_subset}
\end{figure}

\begin{figure}[]
	\centering 
	\includegraphics[width=\linewidth]{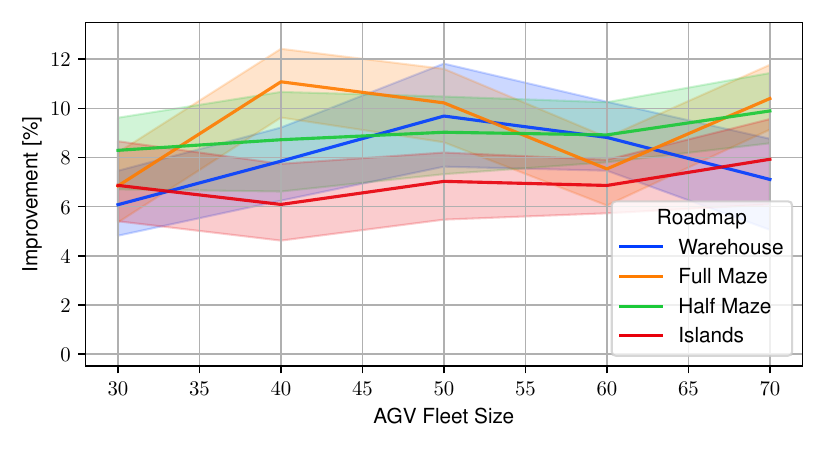}
	\caption{
    Improvement for the roadmaps shown in \figref{fig:roadmaps} 
    for various \acs{AGV} fleet sizes. $20\%$ of the \acs{AGVs}
    are delayed by $\delaytime = 20$s and the $\horizon = 5$s.
    Bounds correspond to $\pm 1 \sigma$.
  }
	\label{fig:diff_maps}
\end{figure}

\subsubsection*{Varying Percentage of Delayed \acs{AGV} Fleet}
We evaluate different proportions of the 
  \acs{AGVs} delayed at each $\delaytime = 20$s,
  with the improvement for an horizon $\horizon = 5$s,
  shown in \figref{fig:eval_improv_vs_subset}.
When $0\%$ of the \acs{AGVs} are delayed,
  the improvement is $0\%$ since none of the 
  \acs{AGVs} are re-ordered with the \acs{RHC} \acs{SADG}
  approach.
Similarly, when $100\%$ of the \acs{AGVs} are delayed,
  the cumulative route completion times for both the 
  \acs{ADG} and \acs{SADG} methods is $\infty$,
  yielding an improvement of $0\%$.
Improvement is highest when $40\%$ randomly selected
  \acs{AGVs}
  are delayed each $\delaytime$.

\subsubsection*{Different Roadmaps}
We evaluate the improvement separately for
  each roadmap in \figref{fig:roadmaps}. 
Results are shown in \figref{fig:diff_maps} 
  for horizon $\horizon = 5$s and 
  $\delaytime = 10$s.
We note that the least improvement is seen for the sparser \textit{Islands} map,
  and the best improvement is seen for the denser \textit{Full Maze} 
  and \textit{Warehouse} maps.
We conclude that our method is best suited to 
  maps which present the \textit{opportunity} for switching,
  as this increases the binary decision space of the \acs{OCP}.

\subsubsection*{\acs{RHC} \acs{MILP} Computation Times}
\begin{figure}[]
  \centering
    \includegraphics[width=0.9\linewidth]{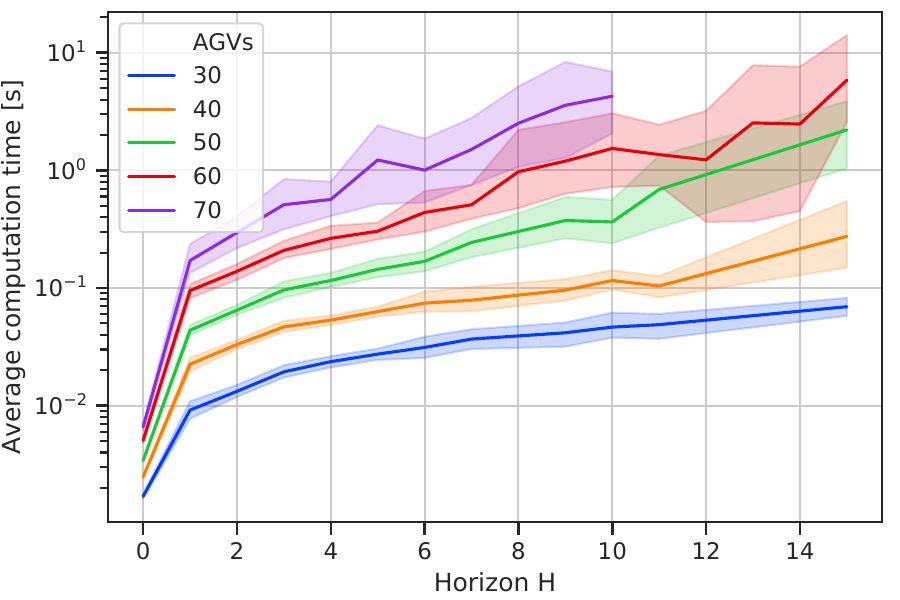}
  \caption{
    Computation times of \textit{3. Optimizer} in \figref{fig:rhc_strategy_diagram} 
    for varying horizon lengths and \acs{AGV} fleet sizes.
    Bounds correspond to $\pm 3 \sigma$.
  }
  \vspace{-2mm}
\label{fig:computation_time}
\end{figure}

For our simulations, 
  the \acs{MILP} was solved using
  the \ac{CBC} solver \cite{johnForrest2018}.
The computation times for different \acs{AGV} fleet sizes and horizon lengths
  are shown in \figref{fig:computation_time}.
We note that the \acs{MILP} can be solved below $1$ second 
  for fleet sizes of up to $70$ \acs{AGVs} for horizons 
  below $\horizon = 5$s.
Recall that even smaller horizons can yield significant improvement 
  as in \figref{fig:horizon_improv}.

\subsubsection*{Summary}
We observe significant reductions in average route completion times 
  for \acs{AGVs}.
Specifically, the larger the delays observed by the \acs{AGVs},
  the larger the improvement is observed when using the \acs{RHC} \acs{SADG}
  approach compared to the \acs{ADG} baseline using the same initial 
  \acs{MAPF} solution.
Significant improvements are observed for small 
  \acs{RHC} \acs{SADG} horizons, across multiple roadmap topologies.

\subsection{Setting 2: Comparison with Robust \acs{MAPF} Solver}
\label{sec:kmapf}

In this section we compare our proposed receding horizon 
  feedback control scheme to the state-of-the-art 
  robust \acs{MAPF} solver \Kalg \cite{ChenAAAI21b}.
\begin{figure}[b]
	\centering
  \begin{subfigure}[h]{0.46\linewidth}
    \centering
    \includegraphics[width=0.85\textwidth]{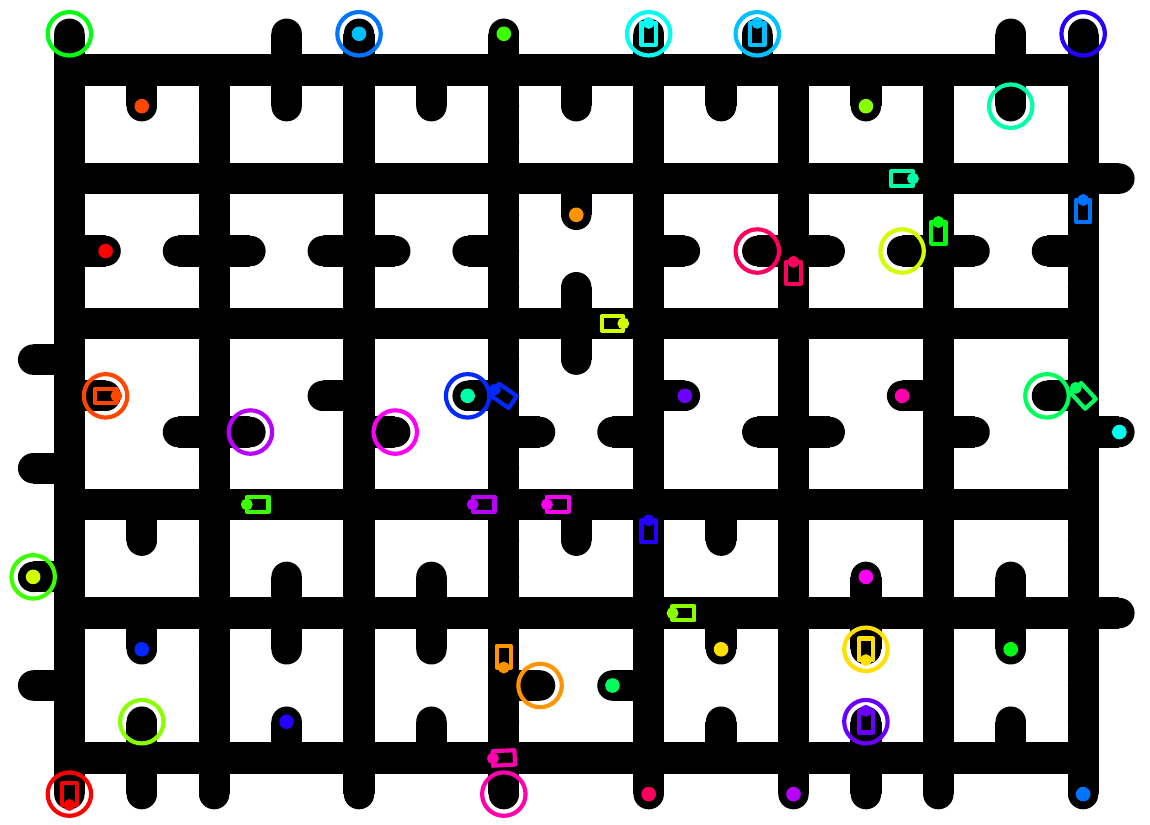}
    \caption{Roadmap for the comparison with \Kalg in \secref{sec:kmapf}}
    \label{fig:kmapf_roadmap}
  \end{subfigure}
  \hspace{1mm}
  \begin{subfigure}[h]{0.46\linewidth}
    \centering
    \includegraphics[width=0.73\textwidth]{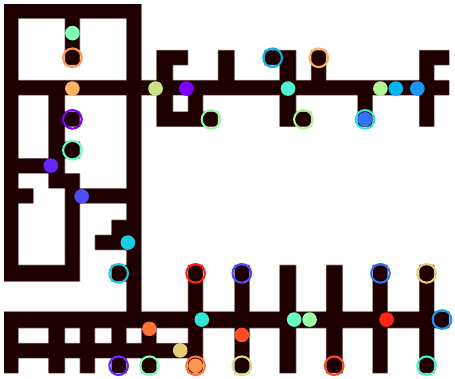}
    \caption{Roadmap used for the Gazebo evaluation in \secref{sec:gazebo_sims}}
    \label{fig:roadmap_gazebo}
  \end{subfigure}
  \caption{Roadmaps with \acs{AGVs} indicated by colored circles,
   and their correspondingly colored rings denoting their goal locations.}
\end{figure}
We consider the roadmap in \figref{fig:kmapf_roadmap}
  with \acs{AGV} fleets of size $20$ and $25$
  with the horizon of the \acs{RHC} \acs{SADG} approach set to $\horizon = 10$s.
The smaller roadmap and fleet sizes were chosen because 
  \Kalg and \acs{CBS} failed to yield valid \acs{MAPF}
  solutions for larger maps or fleet sizes.
Recall that we used the bounded sub-optimal equivalent of \acs{CBS},
  \acs{ECBS}, in Setting 1, 
  allowing us to find feasible solutions 
  to the \acs{MAPF} for larger solution spaces. 
As in Setting 1, 
  improvement is measured using the original \acs{ADG} approach 
  with \acs{CBS} as the initial \acs{MAPF} planner.

\begin{figure}[]
  \centering 
  \includegraphics[width=\linewidth]{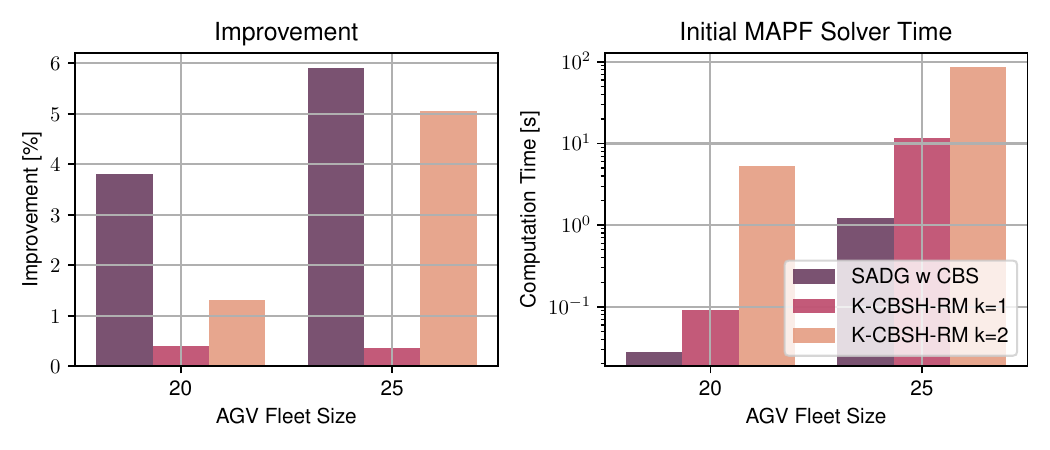}
  \caption{ 
    Average improvement and initial \acs{MAPF} 
    solver computation times for $100$ simulations with \acs{AGVs} 
    velocity profiles from \figref{fig:delay_profiles} comparing 
    the \acs{RHC} \acs{SADG} 
    method with the \Kalg planner for 
    $k \in \{1,2\}$s for $\horizon = 10$s.
  }
  \label{fig:prob_delays}
\end{figure}
\begin{figure}[]
  \centering
  \includegraphics[width=\linewidth]{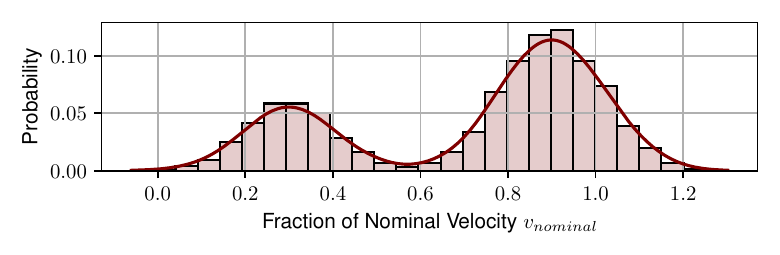}
  \caption{
    Normalized velocity distribution for Setting 2.
    Each \acs{AGV} is given a randomly sampled velocity 
    to complete its next \acs{SADG} event from this distribution. 
  }
  \label{fig:delay_profiles}
\end{figure}
\begin{figure}[]
  \centering
  \includegraphics[width=\linewidth]{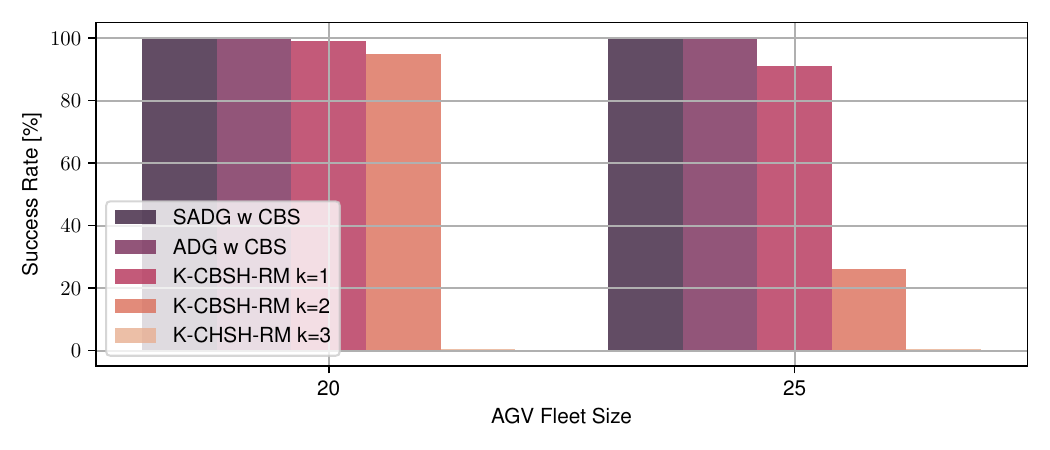}
  \caption{
    Initial \acs{MAPF} planner success for $100$ simulations 
    using \acs{CBS} and \Kalg with $k \in \{1,2,3 \}s$.
  }
  \label{fig:success_rate}
\end{figure}

\subsubsection*{Simulating Interactions with Dynamic Obstacles}
To simulate random interactions with dynamic obstacles, 
  each \acs{AGV} is prescribed a different, randomly sampled 
  velocity when completing an \acs{SE-ADG} event. 
These velocities are sampled 
  from the distribution shown in \figref{fig:delay_profiles}.
This distribution was created to simulate the fact that, 
  most often, \acs{AGVs} move at a velocity close to a nominal velocity 
  $v_\text{nominal}$, but occasionally travel at significantly 
  slower velocities when navigating around dynamic obstacles, 
  modeled here by the velocity distribution centered around
  $0.3v_\text{nominal}$, with $v_\text{nominal} = 1 \ms$.
\figref{fig:prob_delays} shows 
  the comparison of the \acs{RHC} \acs{SADG} approach 
  with \Kalg for $k \in \{1,2\}$ seconds for $100$ simulations.
We observe that the \acs{RHC} \acs{SADG} approach yields a higher improvement
  during plan execution for both \acs{AGV} fleet sizes.
As expected, we observed that improvement increased for increasing values of $k$ 
  in \Kalg. 
However, for $k > 2$, \Kalg only found valid \acs{MAPF} solutions 
  for  $1\%$ of the simulations within the $240$s cut-off time, 
  making a comparison with 
  our approach for $k > 2$ impossible.
The success rate of \Kalg for different values of $k$ 
  is shown in \figref{fig:success_rate}.
These results highlight the fact that our approach 
  can be used with a non-robust planner such as \acs{CBS}
  or \acs{ECBS} to ensure a valid solution to the \acs{MAPF} 
  is found, while adding robustness to delays in an online
  fashion through the re-ordering of \acs{AGV} plans.  

\subsubsection*{Simulating Workspaces with Bounded Delays}
\begin{figure}[]
  \centering 
  \includegraphics[width=\linewidth]{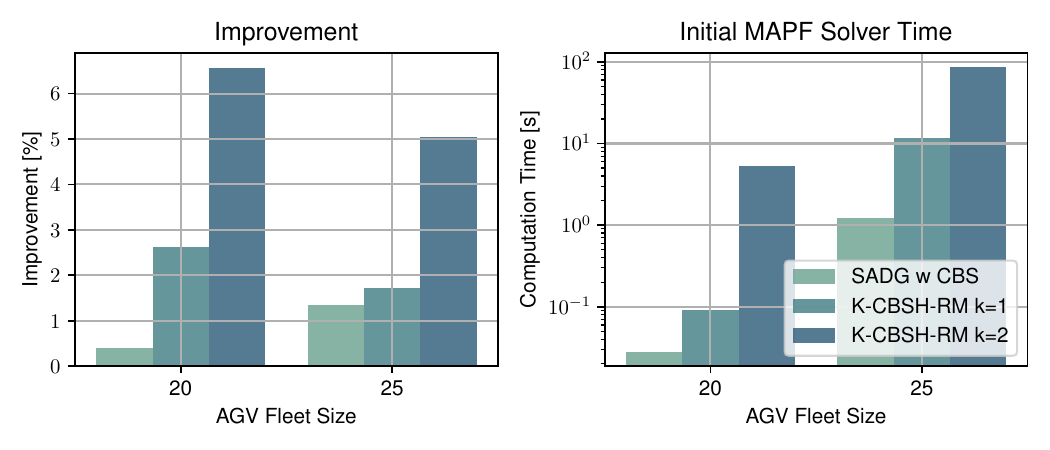}
  \caption{ 
    Average improvement and initial \acs{MAPF} 
    solver computation times for $100$ simulations with \acs{AGVs} 
    velocity profiles from \figref{fig:delay_profiles_2} comparing 
    the \acs{RHC} \acs{SADG}  
    method with the \Kalg planner for 
    $k \in \{1,2\}$s for $\horizon = 10$s.
  }
  \label{fig:prob_delays_2}
\end{figure}

\begin{figure}[]
  \centering
  \includegraphics[width=\linewidth]{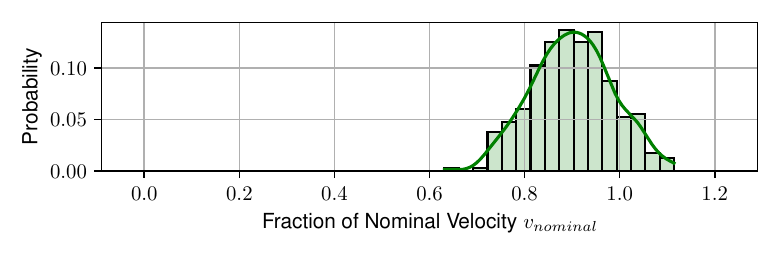}
  \caption{
    Normalized velocity distribution for Setting 2.
    Each \acs{AGV} is given a randomly sampled velocity 
    to complete its next \acs{SADG} event from this distribution. 
  }
  \label{fig:delay_profiles_2}
\end{figure}

The family of robust \acs{MAPF} solvers 
  assume delays of all \acs{AGVs} are upper-bounded 
  by $k$ seconds \cite{atzmonRobustMultiAgentPath2020,ChenAAAI21b}.
We compare our proposed approach to \Kalg 
  with a velocity distribution  
  in \figref{fig:delay_profiles_2} 
  resulting in smaller delays.
\figref{fig:prob_delays_2} shows that \Kalg performs better 
  than our approach in this case, which only yields marginal 
  improvements compared to the \acs{ADG} baseline.

\begin{myremark}
Our approach is agnostic to the original 
  \acs{MAPF} planner. 
Hence,
  it would be possible to use \Kalg to solve the initial 
  \acs{MAPF}, and the \acs{RHC} \acs{SADG} to re-order 
  \acs{AGVs} based on delays observed during plan execution.
\end{myremark}

\subsubsection*{Summary}
We observe that 
  although \Kalg yields theoretically lower 
  cumulative route completion times when delays 
  are bounded by $k$ seconds,
  our proposed \acs{RHC} \acs{SADG} approach
  yields better improvement when delays are larger.
Additionally, the robust \acs{MAPF} planners require
  significantly more time to solve the robust \acs{MAPF} problem,
  as opposed to the standard \acs{MAPF} solver, 
  \acs{CBS} or \acs{ECBS}, used by our approach.


\subsection{Setting 3: High-Fidelity Gazebo Simulations}
\label{sec:gazebo_sims}

We evaluate our proposed approach in a high-fidelity 
  Gazebo simulation environment.
We consider $20$ \acs{AGVs} with randomly selected goal and start locations 
  navigating the roadmap in \figref{fig:roadmap_gazebo}.
\acs{AGVs} use the open-source \acs{ROS} \textit{move\_base} 
  motion planner  
  to execute the \acs{SE-ADG} events.
Delays occur naturally as \acs{AGVs} navigate the workspace and
  negotiate interactions with other dynamic obstacles.
Improvement is quantified as in \eqnref{eq:improvement} 
  and the results are shown in \tabref{tab:comparison_results_3}.
We observe positive improvements for 
  all $20$ simulations with slightly higher 
  improvement in the Gazebo simulation 
  compared to the equivalent start-goal positions 
  when simulating dynamic obstacles in Setting 2.
The Gazebo simulations yielded a higher improvement 
  because \acs{AGVs} were found to experience larger delays 
  than modelled in \figref{fig:delay_profiles}.

\begin{table}[]
	\centering
  \caption{\textsc{Average improvement for $20$ simulations 
  using} \acs{RHC} \acs{SADG} \textsc{with $\horizon = 10$s in Gazebo}}
	\label{tab:comparison_results_3}
  \bgroup
  \def\arraystretch{1.25}
	\begin{tabular}{|l|c|c|}
    \hline
     &
      \textbf{\begin{tabular}[c]{@{}c@{}}Initial MAPF \\ Solver Time {[}s{]}\end{tabular}} &
      \textbf{Improvement {[}\%{]}} \\ \hline \hline
    \begin{tabular}[c]{@{}l@{}}High-Fidelity\\ Gazebo Simulation\end{tabular} &
      3.45 $\pm$ 0.31 &
      8.4  $\pm$ 2.2 \\ \hline
    \begin{tabular}[c]{@{}l@{}}Setting 2: Simulating \\ Interactions with \\ Dynamic Obstacles\end{tabular} &
      3.65  $\pm$ 0.25 &
      6.7 $\pm$ 0.8 \\ \hline
  \end{tabular}
  \egroup
\end{table}

%% file: sec/9_conclusion.tex

\section{Conclusion}
\label{sec:conclusion}

In this manuscript,
  we propose an optimization-based receding horizon 
  feedback control scheme to re-order \acs{AGVs} 
  subject to delays when
  executing \acs{MAPF} plans.
When compared to the state-of-the-art \acs{MAPF} 
  planners,
  our approach yields a significant reduction in 
  cumulative route completion times for \acs{AGVs}
  subjected to large delays, often experienced in 
  uncertain environments with dynamic obstacles.
Our optimization-based re-ordering 
  scheme is derived to obtain approximate solutions  
  to a newly formulated \acf{OCP}. 
This \acs{OCP} is described using a \acf{SADG},
  a novel data-structure introduced in this manuscript.
The \acs{SADG} extends the \acf{ADG} 
  introduced in \cite{hoenigPersistentRobustExecution2018} 
  by enabling the re-ordering 
  of \acs{AGVs} while provably 
  maintaining collision-avoidance guarantees 
  of the original \acs{MAPF} plan.
Moreover, our approach guarantees 
  deadlock-free plan execution 
  while simultaneously minimizing 
  the cumulative route completion time 
  of the \acs{AGVs}.

We evaluate our approach in three settings.
In Setting 1, we
  illustrate the efficiency of our approach,
  reducing the cumulative route completion 
  time for \acs{AGVs} by up to $25\%$
  compared to the baseline \acs{ADG} approach.
Here, we also illustrate the real-time implementability
  of our feedback scheme, 
  showing that the \acs{RHC} \acs{MILP} problem 
  can consistently 
  be solved under one second, even for \acs{AGV}
  fleet sizes of up to $70$ \acs{AGVs},
  all the while significantly reducing route completion times.
In Setting 2,
  we compare our approach to the state-of-the-art in 
  robust \acs{MAPF} planner \Kalg,
  showing a significant reduction in cumulative route completion
  times for \acs{AGVs}
  subjected to larger delays,
  and comparable cumulative route completion times
  for smaller delays.
In Setting 3, we showcased our approach 
  in a high-fidelity Gazebo simulation environment with 
  $20$ \acs{AGVs} navigating around dynamic obstacles,
  reducing the cumulative route completion time by $8\%$,
  thereby validating the results obtained in Settings 1 and 2.

In all simulation settings,
  the \acs{AGVs} exhibited collision- and deadlock-free 
  plan execution,
  a result we prove for both the \acs{SHC} and \acs{RHC}
  feedback schemes.
Our approach is also agnostic
  with respect to the planner used to 
  solve the initial \acs{MAPF} problem.
This means that most \acs{MAPF} planners, 
  e.g., \acs{CBS}, \acs{ECBS}, 
  \Kalg, can be used, 
  as long as the initial \acs{MAPF} solution 
  yields an acyclic \acs{SE-ADG}, 
  a constraint which is easy to adhere to
  as long as the number of roadmap vertices is larger than 
  the number of \acs{AGVs}.
Although we only consider \acs{AGVs} executing intralogistics 
  tasks,
  our approach can be extended to other use-cases 
  covered in the \acs{MAPF} literature.
For future work, we recommend a detailed comparison of
  our approach to real-time re-planning of the 
  \acs{MAPF} using bounded, sub-optimal
  MAPF solvers as in \cite{sternMultiAgentPathfindingDefinitions2019}.

%% file: sec/a_appendix.tex

\appendix

\begin{mylemma}[Two acyclic graphs connected by unidirectional edges 
  yield an acyclic graph]
Consider a directed graph $\mathcal{G} = (\mathcal{V},\mathcal{E})$ subdivided into two subgraphs 
$\mathcal{G}_1 = (\mathcal{V}_1,\mathcal{E}_1)$ and $\mathcal{G}_2 = (\mathcal{V}_2,\mathcal{E}_2)$ such that
$\mathcal{V}_1 \cap \mathcal{V}_2 = \emptyset$ 
and $\mathcal{E}_1 \cap \mathcal{E}_2 = \emptyset$,
$\mathcal{V}_1 \cup \mathcal{V}_2 = \mathcal{V}$
and the edges connecting vertices in $\mathcal{G}_1$ and $\mathcal{G}_2$ are contained 
within the set $\mathcal{E}_{12}$, such that 
$\mathcal{E}_1 \cup \mathcal{E}_2 \cup \mathcal{E}_{12} = \mathcal{E}$. 
If both $\mathcal{G}_1$ and $\mathcal{G}_2$ are acyclic
and $e = (v_1,v_2)$ is such that $v_1 \in \mathcal{V}_1$
and $v_2 \in \mathcal{V}_2$ for all $e \in \mathcal{E}_{12}$,
then the graph $\mathcal{G}$ is also acyclic.
\label{res:two_acyclic_graphs}
\end{mylemma}
\begin{myproof}
Consider two acyclic graphs, $\mathcal{G}_1 = (\mathcal{V}_1, \mathcal{E}_1)$ 
and $\mathcal{G}_2 = (\mathcal{V}_2, \mathcal{E}_2)$.
For $\mathcal{G}_1$, 
consider an inbound edge $e = (v,v')$ which implies that $v \notin \mathcal{V}_1$
and $v' \in \mathcal{V}_1$.
Any number of inbound edges $e$ will not cause $\mathcal{G}_1$ to be cyclic.
Similarly, 
consider an outbound edge $e = (v,v')$ which implies that $v \in \mathcal{V}_1$
and $v' \notin \mathcal{V}_1$.
Any number of outbound edges $e$ will not cause $\mathcal{G}_1$ to be cyclic.
The same arguments apply to $\mathcal{G}_2$.
Since neither $\mathcal{G}_1$ nor $\mathcal{G}_2$ have an internal cycle,
the only possibility for a cycle within $\mathcal{G}$ 
is a cycle through both subgraphs $\mathcal{G}_1$ and $\mathcal{G}_2$.
Since all edges connecting $\mathcal{G}_1$ and $\mathcal{G}_2$ can be defined by
edge $e = (v,v')$ such that $v \in \mathcal{V}_1$ and $v' \in \mathcal{V}_2$,
such a cycle cannot exist.
This guarantees that the entire graph is acyclic,
completing the proof.
\end{myproof}

\begin{figure}[h!]
	\centering
	\includegraphics[width=\linewidth]{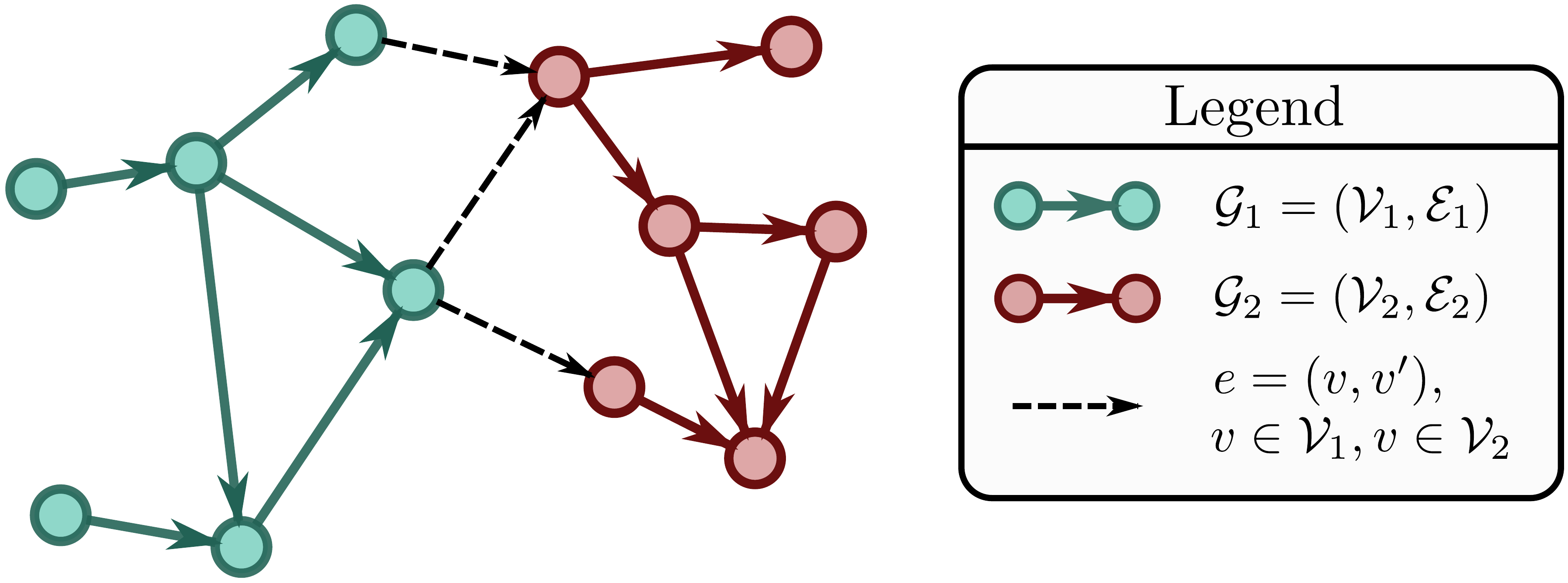}
	\caption{Graphical illustration of \lemref{res:two_acyclic_graphs}: 
	two acyclic graphs $\mathcal{G}_1$ and $\mathcal{G}_2$, connected by uni-directional
	edges, indicated by the dotted lines, yield a larger, acyclic graph.}
	\label{fig:two_acyclic_graphs}
\end{figure} 

%% file: utils/biographies.tex
%

\begin{IEEEbiography}[{\includegraphics[width=1.0in,height=1.25in,clip,keepaspectratio]{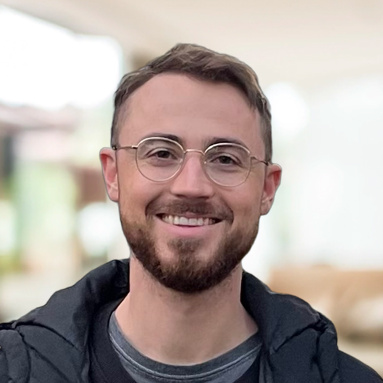}}]{Alexander Berndt}
  is a ML/Software Engineer 
  at Overstory B.V. working on 
  geospatial data intelligence 
  applied to vegetation management.
He received the M.Sc. degree in Systems and Control 
  from the Delft Center for Systems and Control,
  Delft University of Technology, Delft, The Netherlands.
He has authored articles published in 
  ECC, ICAPS and ISTVS spanning the domains of 
  data-driven control, set-based estimation,
  and multi-agent robotics and coordination.
His research interests include 
  data-driven control and estimation schemes
  with guarantees, 
  the control of multi-agent systems,
  and learning-based control.
\end{IEEEbiography}
\vspace{-5mm}
\begin{IEEEbiography}[{\includegraphics[width=1.0in,height=1.25in,clip,keepaspectratio]{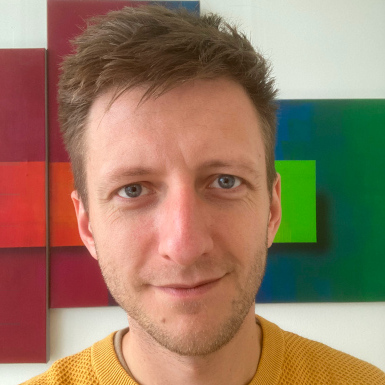}}]{Niels van Duijkeren}
  is a Research Scientist at Robert Bosch GmbH - Corporate Research.
  He received his M.Sc. degree in Systems and Control from Delft University of Technology, The Netherlands
    and his Ph.D. degree from the Motion Estimation Control and Optimization group,
    Mechanical Engineering, KU Leuven, Belgium.
  During his Ph.D., he focused on methods for time-optimal geometric motion control
    of robot manipulators and user-friendly efficient numerical solvers.
  His current research focuses on motion planning and control of mobile robots,
    software and methods for optimization-based control and estimation,
    robust motion planning in dynamic environments,
    system identification and model learning for adaptive robot control.
  He has co-authored papers in e.g. TAC, CDC, IROS, and RSS on topics spanning
    model predictive control, optimization methods, and machine learning.
\end{IEEEbiography}
\vspace{-5mm}
\begin{IEEEbiography}[{\includegraphics[width=1.0in,height=1.25in,clip,keepaspectratio]{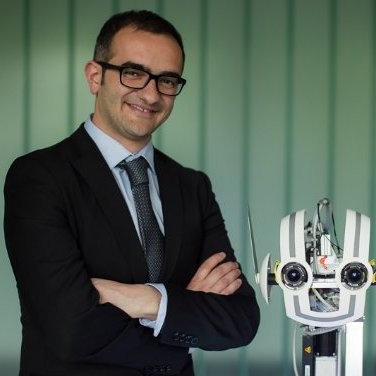}}]{Luigi Palmieri}
is a Senior Expert at Robert Bosch GmbH - Corporate Research. 
His research currently focuses on kinodynamic 
  motion planning in dynamic and crowded environments, 
  control of non linear dynamic systems, 
  hybrid systems of learning-planning-control, 
  MPC and numerical optimization techniques, 
  planning considering human motion predictions and social constraints.
He earned his Ph.D. degree in robot motion planning 
  from the University of Freiburg, Germany.
During his Ph.D., he was responsible for 
  the motion planning task of the EU FP7 project Spencer.
Since then, he has the same responsibility 
  in the EU H2020 project ILIAD. 
He has co-authored multiple papers at RA-L, ICRA, IROS, 
  FSR on the combinations of motion planning with control,
  search, machine learning, and human motion prediction.
\end{IEEEbiography}
\vspace{-5mm}
\begin{IEEEbiography}[{\includegraphics[width=1.0in,height=1.25in,clip,keepaspectratio]{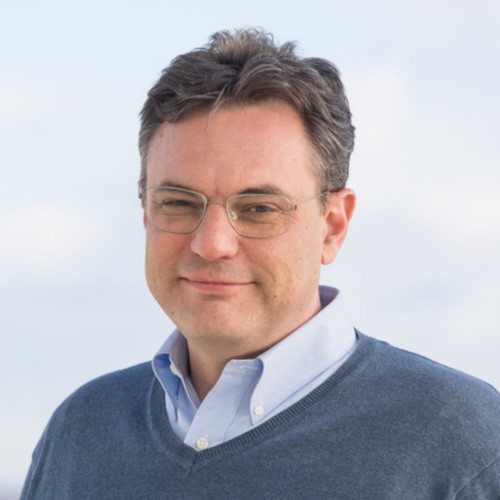}}]{Alexander Kleiner}
  is chief expert for navigation and coordination
    of autonomous systems at Bosch Cooperate Research.
  From 2017 until 2018 he worked as President for AI and 
    Machine Learning at the startup FaceMap LLC in Malibu, CA and from 2014 
    until 2018 as Senior Principal Robotics Scientist 
    with technical lead at iRobot in Pasadena, CA. 
  From 2011 to 2014 he served as associate professor at 
    the Link{\"o}ping University in Sweden where he 
    headed the research group on collaborative robotics. 
  He holds a M.Sc. degree in computer science from the 
    Stafford University in UK, a Ph.D. degree in computer science 
    from the University of Freiburg in Germany, and a 
    docent degree (habilitation) from the Link{\"o}ping University in Sweden. 
  He worked as a postdoctoral fellow at Carnegie 
    Mellon University in the US and at La Sapienza University in Italy. 
  His research interests include collaborative robotics, 
    multi-robot navigation planning, and machine learning.
\end{IEEEbiography}
\begin{IEEEbiography}[{\includegraphics[width=1.0in,height=1.25in,clip,keepaspectratio]{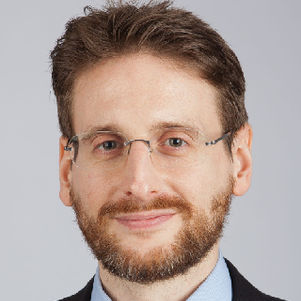}}]{Tam{\'a}s Keviczky}
(Senior Member, IEEE) received the M.Sc. degree in
  electrical engineering from Budapest University of Technology and
  Economics, Budapest, Hungary, in 2001, and the Ph.D. degree from the
  Control Science and Dynamical Systems Center, University of Minnesota,
  Minneapolis, MN, USA, in 2005.
He was a Post-Doctoral Scholar of control and dynamical systems with
  California Institute of Technology, Pasadena, CA, USA. 
He is currently a
  Professor with Delft Center for Systems and Control, Delft University of
  Technology, Delft, The Netherlands. 
His research interests include
  distributed optimization and optimal control, model predictive control,
  embedded optimization-based control and estimation of large-scale
  systems with applications in aerospace, automotive, mobile robotics,
  industrial processes, and infrastructure systems, such as water, heat,
  and power networks.
He was a co-recipient of the AACC O. Hugo Schuck Best Paper Award for
  Practice in 2005. 
He has served as an Associate Editor for Automatica
  from 2011 to 2017 and for IEEE Transactions on Automatic Control since 2021. 
\end{IEEEbiography}


